\documentclass[fleqn,12pt]{article}

\usepackage{times}
\usepackage{url}
\usepackage[utf8]{inputenc}
\usepackage[small]{caption}
\usepackage{graphicx}
\usepackage{amsmath}
\usepackage{amssymb}
\usepackage{amsthm}
\usepackage{booktabs}
\usepackage[ruled]{algorithm2e}
\usepackage{algorithmic}
\usepackage{ifthen}

\usepackage{xspace}
\usepackage{color}
\usepackage{enumitem}

\newtheorem{definition}{Definition}[section]
\newtheorem{example}[definition]{Example}
\newtheorem{theorem}[definition]{Theorem}
\newtheorem{proposition}[definition]{Proposition}
\newtheorem{lemma}[definition]{Lemma}

\newtheorem{remark}[definition]{Remark}

\setlist{topsep=3pt, itemsep=-3pt} 

\SetAlCapSkip{1mm}
\SetAlFnt{\small\rm}
\SetAlCapFnt{\small\rm}

\newenvironment{program}[1][htb]
{
 \begin{algorithm}[#1]
 }{\end{algorithm}%
}
\LinesNumbered
\DontPrintSemicolon
\SetAlgoLined

\newcommand{\done}{\hfill\ensuremath{\Box}}
\newcommand{\tuple}[1]{\ensuremath{\langle #1\rangle}\xspace}
\renewcommand{\implies}{\rightarrow}
\newcommand{\problog}{\textsc{ProbLog}\xspace}
\newcommand{\prolog}{\textsc{Prolog}\xspace}
\newcommand{\dls}{\textsc{Dls}\xspace}
\newcommand{\when}{\mbox{\,:--\;}\xspace}

\newcommand{\calc}{\ensuremath{\mathcal{C}}\xspace}
\newcommand{\cale}{\ensuremath{\mathcal{E}}\xspace}
\newcommand{\calf}{\ensuremath{\mathcal{F}}\xspace}
\newcommand{\calg}{\ensuremath{\mathcal{G}}\xspace}
\newcommand{\calm}{\ensuremath{\mathcal{M}}\xspace}
\newcommand{\calp}{\ensuremath{\mathcal{P}}\xspace}
\newcommand{\calr}{\ensuremath{\mathcal{R}}\xspace}
\newcommand{\calt}{\ensuremath{\mathcal{T}}\xspace}
\newcommand{\calv}{\ensuremath{\mathcal{V}}\xspace}
\newcommand{\calw}{\ensuremath{\mathcal{W}}\xspace}

\newcommand{\defeq}{\stackrel{\mathrm{def}}{=}}
\newcommand{\defequiv}{\stackrel{\mathrm{def}}{\equiv}}

\newcommand{\lneg}{\neg}
\newcommand{\pneg}{\mbox{\scriptsize$\backslash$+\,}\xspace}

\newcommand{\true}{\ensuremath{\mathbb{T}}\xspace}
\newcommand{\false}{\ensuremath{\mathbb{F}}\xspace}

\newcommand{\nforget}[1]{\ensuremath{F^{NC}(#1)}\xspace}
\newcommand{\sforget}[1]{\ensuremath{F^{SC}(#1)}\xspace}

\newcommand{\prob}[1]{\ensuremath{\calp\big(#1\big)}\xspace}
\newcommand{\probw}[1]{\ensuremath{\calp_\calw\big(#1\big)}\xspace}
\newcommand{\probz}[1]{\ensuremath{\calp_0\big(#1\big)}\xspace}

\newcommand{\nlossm}[1]{\ensuremath{loss_{m}^{NC}\big(#1\big)}\xspace}
\newcommand{\slossm}[1]{\ensuremath{loss_{m}^{SC}\big(#1\big)}\xspace}
\newcommand{\tlossm}[1]{\ensuremath{loss_{m}^{T}\big(#1\big)}\xspace}
\newcommand{\lossm}[1]{\ensuremath{loss_{m}^{*}\big(#1\big)}\xspace}

\newcommand{\nlossp}[1]{\ensuremath{loss_{p}^{NC}\big(#1\big)}\xspace}
\newcommand{\slossp}[1]{\ensuremath{loss_{p}^{SC}\big(#1\big)}\xspace}
\newcommand{\tlossp}[1]{\ensuremath{loss_{p}^{T}\big(#1\big)}\xspace}
\newcommand{\lossp}[1]{\ensuremath{loss_{p}^{*}\big(#1\big)}\xspace}

\title{Techniques for Measuring the Inferential Strength of~Forgetting Policies}

\author{ Patrick Doherty$^{1,3}$,
Andrzej Sza{\l}as$^{1,2}$ \\
$^1$Department of Computer and Information Science,\\
  Link\"{o}ping University, Sweden\\
$^2$Institute of Informatics, University of Warsaw, Poland \\
$^3$ Faculty of Informatics, Mahasarakham University, Thailand
\\[0.5em]
patrick.doherty@liu.se,
andrzej.szalas@\{mimuw.edu.pl, liu.se\}
}

\date{
}

\begin{document}

\maketitle

\begin{abstract}
The technique of \emph{forgetting} in knowledge representation has been shown to be a powerful and useful knowledge engineering tool with widespread application. Yet, very little research has been done on how different policies of forgetting, or use of different forgetting operators, affects the inferential strength of the original theory. The goal of this paper is to define loss functions for measuring changes in inferential strength based on intuitions from model counting and probability theory. Properties of such loss measures are studied and a pragmatic knowledge engineering tool is proposed for computing loss measures using a~well-known probabilistic logic programming language \problog. 
The paper includes a working methodology for studying and determining the strength of different forgetting policies, in addition to concrete examples showing how to apply the theoretical results using \problog. Although the focus is on forgetting, the results are much more general and should have wider application to other areas.
\end{abstract}

\section{Introduction}\label{sec:intro}

The technique of forgetting~\cite{Lin94forgetit} in knowledge representation has been shown to be a powerful and useful knowledge engineering tool with many applications (see Section~\ref{sec:relwork} on related work). 
In~\cite{dsz-forgetting-aij}, it has been shown that  two symmetric and well-behaved forgetting operators can be defined that provide a qualitative \emph{best} upper and lower bound on forgetting in the following sense: it has been shown that strong or standard forgetting~\cite{Lin94forgetit} provides a strongest necessary condition on a~theory with a specific forgetting policy, while weak forgetting~\cite{dsz-forgetting-aij} provides a weakest sufficient condition on a theory with a specific forgetting policy. In the former, one loses inferential strength relative to the original theory wrt necessary conditions, while in the latter one 
loses inferential strength wrt sufficient conditions. That is, some necessary (respectively, sufficient) conditions that can be expressed in the language of the original theory may be lost when the language becomes restricted after forgetting. 
A question then arises as to how these losses can be measured, both individually relative to each operator and comparatively relative to each other. The loss function used should be well-behaved and also quantitative, so as to ensure fine-grained measurement.

The goal of this paper is to define such  measures in terms of loss functions, show useful properties associated with them, show how they can be applied, and also show how they can be computed efficiently. The basis for doing this will be to use intuitions from the area of model counting, both classical and approximate~\cite{fichte,GomesSS21} and probability theory. 
In doing this, a~pragmatic knowledge engineering tool based on the use of a~well-known probabilistic logic programming language \problog~\cite{RaedtK15} will be described that allows for succinctly identifying the inferential strength of arbitrary theories, in themselves, and their associated losses using the dual forgetting operators. Initial focus is placed on propositional theories for clarity, but the techniques are then shown to generalize naturally to 1st-order theories, with some restrictions. 

Although the focus in this paper is on measuring inferential strength relative to forgetting operators, the techniques are much more general than this and should be applicable to other application areas including traditional model counting, approximate model counting through sampling, and theory abstraction.

Let's begin with describing an approach to measuring inferential strength using intuitions from model counting and then show how this measurement technique can be used in the context of forgetting.
Consider the following simple theory, $\calt_c$, extracted from an example considered in~\cite{NayakL95} which deals with theory abstraction, where $jcar$ stands for ``Japanese car'', $ecar$ -- for ``European car'', and $fcar$ -- for ``foreign car'':
\begin{align}
    & jcar\implies (car\land reliable\land fcar)\label{eq:cars1}\\[-0.3em]
    & ecar\implies (car\land fast\land  fcar)\label{eq:cars2}\\[-0.3em]
    & fcar\implies (ecar\lor jcar). \label{eq:cars3}
\end{align}

\noindent The theory partitions the set of all assignments of truth values to propositional variables occurring in a theory ({\em worlds}) into two sets: (i) the worlds satisfying $\calt_c$ ({\em models}), and (ii) the worlds not satisfying $\calt_c$. This suggests a use of model counting where one measures the proportion among models that satisfy and do not satisfy the theory. Though model counting enjoys quite successful implementations~\cite{fichte}, it is generally  complex~\cite{GomesSS21} with challenging scalability issues. However, in a more general case, the number of models satisfying a theory $\calt$ is proportional to the {\em probability of $\calt$}, another useful measure of inferential strength. Here we shall deal with probability structures, where samples are worlds, events are sets of worlds specified by formulas (given as models of formulas). Since we deal with finite sets of worlds only, the probability function is given by,
\begin{equation}\label{eq:probth}
    \prob{\calt}\defeq
    \frac{|\{w\in\calw: w\models \calt\}|}{|\calw|},
\end{equation}
where \calt is a theory, $\calw$ is the set of worlds, $\models$ is the satisfiability relation, and $|.|$ denotes the cardinality of a set. That is, $ \prob{\calt}$ is the fraction between the number of models of \calt and the number of all worlds.\footnote{Note that given the probability of a theory $\calt$ and the number of variables in the theory, $n$, the model count can easily be derived using~\eqref{eq:probth}, being in this case $\prob{\calt}*|\calw|=\prob{\calt}*2^n$.} 

For example, there are $2^6\!=\!64$ assignments of truth values to propositional variables occurring in $\calt_c$ (worlds) and  $13$ such assignments satisfying $\calt_c$ (models of $\calt_c$), so $\prob{\calt_c}=13/64=0.203125$. 
Since evaluating probabilities can, in many cases, be very efficient (e.g., approximated by sampling~\cite{RaedtK15}), focus will be placed on probabilities rather than model counting directly. More general probabilistic loss measures will be defined, where one does not just count models with uniform distributions, but also pays attention to arbitrary probability distributions. This provides a much more flexible and general inferential strength measurement technique. Note also that the use of probabilities rather than (weighted) model counting results in measures normalized to the interval $[0.0, 1.0]$ what makes them relatively easy to interpret.

These intuitions provide the basis for the specification of  loss functions for forgetting operators, based on probability distributions on worlds, that measure the inferential strength of the original theory and the loss 
in inferential strength after applying a specific forgetting policy and operator. The term {\em forgetting policy}, when used in this paper, refers to the selection of symbols to forget and the choice of the forgetting operator. However, in applications one can develop more advanced policies, where the choice is context-dependent, e.g., based on system states, environmental conditions, etc. In such cases, the loss functions we propose can still be applied to assess each final choice of symbols to forget and which operators to use, which gives the policy designers a~tool to compare policies and gain a much better view on the resulting inferential strength under specific conditions.

Two forgetting operators considered in \cite{dsz-forgetting-aij}, strong (standard) and weak forgetting, are shown to have very nice intrinsic properties in the following sense. Given a specific policy of forgetting, when applying that policy to a theory using the weak forgetting operator, the resultant models of the revised theory are a subset of models of the original theory. Similarly, when applying the strong forgetting operator to the same theory using the same forgetting policy, the resultant models of the revised theory are a superset of the models of the original theory.  As mentioned previously, the revised theories characterize the weakest sufficient and strongest necessary theories relative to a specific forgetting policy.

\label{page:infstrength} When particular symbols are forgotten from a theory, its inferential strength decreases. In fact, the set of models of the original theory is included in (when strong forgetting is concerned) or includes (in the case of weak forgetting) the set of models after forgetting.\footnote{See Equation~\eqref{eq:inclusion} later in the current paper.}  This suggests loss measures based on model counting to reflect losses in inferential strength related to forgetting. Namely, given theories $\calt$ and $\calt'$ obtained from \calt by forgetting some symbols, we have the following properties, where {\em more} and {\em less} are understood wrt  $\subseteq$:
\begin{itemize}
	\item the more models   $\calt'$ has, the less conclusions (necessary conditions) it may entail compared to the original theory $\calt$;\label{page:bulleted};
	\item    the less  models $\calt'$ has, the less sufficient conditions it is entailed by comparing to the original theory $\calt$.
\end{itemize}  

One can therefore measure the gap between the original theory and the theories resulting from forgetting in terms of the size of gaps between the mentioned sets of models associated with the revised theories. Intuitively, the larger the gaps are, the more one loses in terms of inferential strength.

The motivations for desiring such measures include the following:\label{page:motivations}
\begin{itemize}
    \item no measures have yet been provided in the literature that indicate how inferential strength changes when the technique of forgetting is used;
    \item such measures are useful indicators for belief base engineers when determining proper policies of forgetting for various applications;
    \item such measures are generally needed as one of the criteria that can be used for automated selection of symbols to forget.
\end{itemize}

\noindent  Among numerous applications of forgetting, where quantitative measures can serve as a useful tool, let us emphasize the following ones:
\begin{itemize}
	\item optimizing a selected class of queries, when forgetting a particular choice of symbols can substantially increase the system's performance without seriously affecting its inferential capabilities;
	\item information hiding, when one is interested in decreasing inferential capabilities wrt given symbols as much as possible;
	\item abstracting, when one selects details to forget about while keeping the resultant theories sufficiently strong;
    \item knowledge compilation and theory approximation, when one looks for less complex, typically tractable, theories that bound the original theory from below and above, while preserving as much as possible in terms of inferential strength. \label{page:applications}
\end{itemize}

The working methodology for the approach is as follows, where details are provided in the paper:
\begin{enumerate}
    \item[1.] Given a propositional theory $\calt$, transform it into a~stratified logic program using the transformation rules in Section~\ref{sec:compmodcount}.
    \item[2.] Specify a probability distribution on worlds using the propositional facts (variables in $\calt$) as probabilistic variables in \problog, as described in Section~\ref{sec:problog}. 
    \item[3.] Using \problog's query mechanism, described in Section~\ref{sec:problog}, query for the probability $\prob{\calt}$ of theory $\calt$.
\end{enumerate}
For forgetting, given a propositional theory $Th(\bar{p},\bar{q})$ and a~forgetting policy $\bar{p}$:
\begin{enumerate}
    \item[4.] Specify the 2nd-order theories, $\nforget{Th(\bar{p},\bar{q}); \bar{p}}$ and $\sforget{Th(\bar{p},\bar{q}); \bar{p}}$, representing strong and weak forgetting, respectively, relative to the forgetting policy $\bar{p}$, as described in Section~\ref{sec:forgetting}.
    \item[5.] Apply 2nd-order quantifier elimination techniques~\cite{dls,dsz-forgetting-aij,gss} to both theories as exemplified in Section~\ref{sec:example-prop}, resulting in propositional theories without $\bar{p}$.
    \item[6.] Apply steps 1 to 3 above to each of the theories.
    \item[7.] The resulting probabilities $\prob{\calt}$, $\prob{\nforget{\calt;\bar{p}}}$, and $\prob{\sforget{\calt;\bar{p}}}$,  for each of the three theories, can then be used to compute \emph{loss} values for the respective  losses of inferential strength associated with strong and weak forgetting relative to the original theory, as described in Section~\ref{sec:compmodcount} and Section~\ref{sec:compprob}. 
\end{enumerate}
Generalization of this approach to the 1st-order case is described in Section~\ref{sec:fo}.

The original results of this paper include:
\begin{itemize}
    \item formal specification of model counting-based  and probability-based loss measures for forgetting with an analysis of their properties;
    \item an equivalence preserving technique for transforming any formula of propositional or 1st-order logic into a set of stratified logic programming rules in time linear wrt the size of the input formula;
    \item algorithms for computing these loss measures implemented using \problog.
\end{itemize}
\problog is chosen as a basis for this research since it provides a straightforward means of computing probabilistic measures in general, and it also has built in mechanisms for doing approximate inference by sampling~\cite{RaedtK15} (constant time by fixing the number of samples). Consequently, both classical and approximation-based model counting techniques can be leveraged in defining loss measures.

The paper is structured as follows. In Section~\ref{sec:prelim}, the logical background, essentials of strong and weak forgetting and the \problog constructs used in the paper are described. In Section~\ref{sec:modcountmeas}, model counting-based loss measures for forgetting are defined, their  properties are shown and a technique is provided  for computing such measures using \problog. In Section~\ref{sec:probmeas}, 
probabilistic-based loss measures are defined which allow for arbitrary probability distributions over worlds rather than uniform distributions. It is also shown how one computes probabilistic-based loss measures using \problog. In Section~\ref{sec:example-prop}, an example is considered showing how both types of loss measures are used  to assess loss of inferential strength relative to forgetting policies. 
Section~\ref{sec:fo} provides a generalization of the techniques to the 1st-order case. The generalized techniques are exemplified in Section~\ref{sec:example-fo}. Section~\ref{sec:relwork} is devoted to a~discussion of related work.  Section~\ref{sec:concl} concludes the paper with summary and final remarks. Finally, \problog sources of all programs discussed in the paper are listed in the Appendix in a ready to validate form using an online \problog interpreter or its standalone version referenced in Section~\ref{sec:problog}.

\section{Preliminaries}\label{sec:prelim}

\subsection{Logical Background}\label{sec:logic-background}

For the sake of simplicity, the major ideas will be presented by focusing on classical propositional logic with truth values \true (true) and \false (false), an enumerable set of propositional variables, $\calv^0$, and standard connectives $\lneg,\land,\lor,\implies,\equiv$. It is assumed that the classical two-valued semantics is used with truth values \true (true) and \false (false). The set of propositional formulas is denoted by $\calf^0$.

A {\em theory} is a finite set of propositional formulas.\footnote{Since we do not assume that theories are deductively closed, such finite sets of formulas are frequently called {\em belief bases}.\label{footnote:dedclosed}} As usual, propositional theories are identified with conjunctions of their formulas. By a {\em vocabulary} (a {\em signature}) of \calt, denoted by $\calv^0_\calt$, one means all propositional variables occurring in \calt.

By a {\em world for a theory (formula)} \calt, one means any assignment of truth values to propositional variables in $\calv^0_\calt$:
\begin{equation}\label{eq:world-propositional}
    w: \calv^0_\calt\longrightarrow\{\true, \false\}.
\end{equation}
The set of worlds of \calt is denoted by $\calw_\calt$. Note that the considered theories always contain a~finite set of variables, so the number of worlds is finite. The definition~\eqref{eq:world-propositional} can be extended to all propositional formulas in the standard way. For $A\in\calf^0$, by $w\models A$ we mean that $w(A)=\true$. For $A, B\in\calf^0$, we say that $B$ is a {\em consequence of} $A$ ($B$ is {\em entailed by} $A$), denoted by $A\models B$, when for every world $w$, $w\models A$ implies $w\models B$.\footnote{By the deduction theorem for propositional logic, we have that $A\models B$ iff $\models A\implies B$.} Writing $\models A$ we mean that $w\models A$ for every world $w$ (i.e., $A$ is a tautology).

We say that a formula $A$ is {\em at least as strong as} a formula $B$ (wrt $\implies$), if $A\implies B$ is a~tautology ($\models A\implies B$). In such a case, we also say that $B$ is {\em at least as weak as} $A$ (wrt $\implies$), $A$ is a~{\em sufficient condition} for $B$, and $B$ is a {\em necessary condition} for $A$. 
A formula  $A$ is {\em strongest} (wrt $\implies$), satisfying a given property $\mathbb{P}$ if $A$ satisfies $\mathbb{P}$ and for every formula $B$ satisfying $\mathbb{P}$, we have that $A\implies B$ is a tautology. A formula $A$ is {\em weakest} (wrt $\implies$), satisfying a given property $\mathbb{P}$ if $A$ satisfies $\mathbb{P}$ and for every formula $B$ satisfying $\mathbb{P}$, we have that $B\implies A$ is a tautology. 

A formula (theory)  $A$ is  {\em inferentially stronger than or equal to} formula (theory)  $B$ {\em wrt necessary conditions} if the set of consequences of $A$ includes the set of consequences of $B$, i.e., $\{C\mid A\models C\}\supseteq \{C\mid B\models C\}$.
Formula $A$ is {\em inferentially weaker than or equal to} $B$ {\em wrt necessary conditions} when the set of consequences of $A$ is included in the set of consequences of $B$,  $\{C\mid A\models C\}\subseteq \{C\mid B\models C\}$.

The inference {\em by} sufficient conditions and the usual inference {\em of} necessary conditions work in opposite  directions. Therefore we also have a dual definition of inferential strength: a formula (theory)  $A$ is  {\em inferentially stronger than or equal to}  formula (theory)  $B$ {\em wrt sufficient conditions} if $\{C\mid C\models A\}\supseteq \{C\mid C\models B\}$.
Formula $A$ is {\em inferentially weaker than or equal to} $B$ {\em wrt sufficient conditions} when $\{C\mid C\models A\}\subseteq \{C\mid C\models B\}$.

We have the following proposition, where $mod(A)$ denotes the set of models of $A$.

\begin{proposition}\label{prop:strength}
     Let $A,B\in\calf^0$ satisfy $mod(A)\subseteq mod(B)$. Then:
     \begin{itemize}
         \item $A$ is inferentially stronger or equal to $B$ wrt necessary conditions, and $B$ is inferentially weaker or equal to $A$ wrt necessary conditions. 
         \item $A$ is inferentially weaker or equal to $B$ wrt sufficient conditions, and $B$ is inferentially stronger or equal to $A$ wrt necessary conditions. \done
     \end{itemize}
\end{proposition}

When it is clear from the context, we sometimes refer to  {\em inferential strength} without explicitly specifying whether we refer to the strength wrt necessary or sufficient conditions.

\begin{remark}\label{rem:ncscmodels}
Note that when one approximates a theory \calt from below  and above, say by theories $\calt_l$ and $\calt_u$, respectively, i.e., $\models \calt_l\implies\calt$ and $\models\calt\implies\calt_u$, as will happen in the case of forgetting operators (see~\eqref{eq:forgetimplies}), then:
\begin{itemize}
    \item  the less models that satisfy the upper approximation $\calt_u$, the inferentially stronger or equal (wrt necessary conditions) the approximation is, since the set of consequences (necessary conditions) entailed by $\calt_u$ can only grow (or remain the same); 
    \item the more models that satisfy the lower approximation $\calt_l$, the inferentially stronger or equal (wrt sufficient conditions) the approximation is, since the set of sufficient conditions that entail the approximation $\
    \calt_l$  can only grow  (or remain the same).\done
\end{itemize}
\end{remark}

To define forgetting operators, 2nd-order quantifiers $\exists p, \forall p$, will be used where $p$ is a~propositional variable. The meaning of quantifiers in the propositional context is:
\begin{align}
		&\exists p\big(A(p)\big)\defequiv A(p=\false)\lor A(p=\true);\label{eq:exists}\\[-0.5em]
		&\forall p\big(A(p)\big)\defequiv A(p=\false)\land A(p=\true),\label{eq:forall}
\end{align} 
where, for a propositional formula $B$, $A(p=B)$ denotes the formula obtained from $A$ by substituting all occurrences of propositional variable $p$ in $A$ by  $B$.

Finally, by the {\em size of a formula} we understand its length in terms of the number of variables and propositional connectives it contains.

\subsection{Forgetting}\label{sec:forgetting}

This subsection is based on~\cite{dsz-forgetting-aij}. Let $\calt(\bar{p},\bar{q})$ be a propositional theory over a~vocabulary consisting of propositional variables in tuples $\bar{p}, \bar{q}$.\footnote{When one refers to tuples of variables as arguments, as in $Th(\bar{p}, \bar{q})$, it is assumed that $\bar{p}$ and $\bar{q}$ are disjoint.} When forgetting $\bar{p}$ in the theory $Th(\bar{p},\bar{q})$, one can delineate two alternative views, with both resulting in a theory expressed in the vocabulary containing $\bar{q}$ only:
	\begin{itemize}
		\item {\em strong (standard) forgetting} $\nforget{Th(\bar{p},\bar{q}); \bar{p}}$: the (strongest wrt $\implies$) theory that preserves the entailment of necessary conditions over vocabulary $\bar{q}$; that is, for any formula $A$ on a~vocabulary disjoint with $\bar{p}$, $\models Th(\bar{p},\bar{q})\implies A \mbox{ iff } \models \nforget{Th(\bar{p},\bar{q});\bar{p}}\implies A$;
		\item {\em weak forgetting} $\sforget{Th(\bar{p},\bar{q}); \bar{p}}$: the (weakest wrt $\implies$) theory that preserves the entailment by sufficient conditions over $\bar{q}$; that is, for any formula $A$ on a vocabulary disjoint with $\bar{p}$, $\models A\implies Th(\bar{p},\bar{q}) \mbox{ iff } \models A\implies\sforget{Th(\bar{p},\bar{q});\bar{p}}$.
	\end{itemize}

Forgetting and dual forgetting can be seen as strongest necessary and weakest sufficient conditions of a given theory~\cite{dsz-forgetting-aij} expressed in its sublanguage. 

\noindent As shown in~\cite{dsz-forgetting-aij}, the following theorem holds.

	\begin{theorem}\label{thm:forget}
		For arbitrary tuples of propositional variables $\bar{p}, \bar{q}$ and $Th(\bar{p},\bar{q})$,
		\begin{enumerate}
			\item 
			$\nforget{Th(\bar{p},\bar{q}); \bar{p}}  \equiv \exists \bar{p}\big(Th(\bar{p},\bar{q})\big).$
    \item 			$    \sforget{Th(\bar{p},\bar{q}); \bar{p}}\equiv \forall \bar{p}\,\big(Th(\bar{p},\bar{q})\big).
			$
   \done
		\end{enumerate}
	\end{theorem}

\noindent From Theorem~\ref{thm:forget} it easily follows that:
\begin{equation}\label{eq:forgetimplies}
\begin{array}{l}
     \models \sforget{Th(\bar{p},\bar{q}); \bar{p}}\implies Th(\bar{p},\bar{q})\mbox{ and }
     \models Th(\bar{p},\bar{q}) \implies \nforget{Th(\bar{p},\bar{q}); \bar{p}}.
\end{array}
\end{equation}

\subsection{ProbLog}\label{sec:problog}

Let us start with a definition of the {\em probability of a theory}. In the paper we follow the standard approach~\cite{fhm,hal-book}. 

\begin{definition}\label{def:probtheory}
    Let a theory \calt be given. By a {\em probability structure} over the vocabulary of \calt, $\calv^0_{\calt}$, we understand the tuple $\calm\defeq\tuple{\calw^\calm, \cale^\calm, \calp^{\calm}}$, where:
\begin{itemize}
    \item $\calw^\calm\defeq \calw_\calt$  is the  set of all worlds over $\calv^0_{\calt}$  (the set of {\em samples}); 
    \item  $\cale^\calm\defeq \{E\mid E\subseteq \calw^\calm\}$ is the set of {\em events}, each event being a subset of $\calw^\calm$;
    \item $\calp^\calm:\cale^\calm\longrightarrow [0.0, 1.0]$ is a {\em probability function} assigning probabilities to events.
\end{itemize}
\noindent We define $\calt^\calm$ to be the set of worlds where $\calt$ is true, $\calt^\calm\defeq\{w\in \calw^\calm\mid w\models \calt \}$. The {\em probability of theory $\calt$ wrt \calm} is defined as $\calp^\calm(\calt^\calm)$.\done 
\end{definition}

\begin{remark}\label{rem:probability}
   Probability structures allow for arbitrary probability distributions over the set of worlds. When the distribution is uniform, with all worlds having the same probability, then the probability function is given by~\eqref{eq:probth}. Other distributions are also allowed, as we discuss in Section~\ref{sec:probmeas}.   \done
\end{remark}

From~\eqref{eq:forgetimplies} the following important property holds, where  $\prob{}$ is an arbitrary probability function in the sense of Definition~\ref{def:probtheory}:
\begin{equation}\label{eq:important}
     \prob{\sforget{Th(\bar{p},\bar{q}); \bar{p}}}\leq\prob{Th(\bar{p},\bar{q})}\leq \prob{ \nforget{Th(\bar{p},\bar{q}); \bar{p}}}.
\end{equation}

To compute the probability of \calt, one can use the  probabilistic logic programming language \problog~\cite{RaedtK15}.
\problog extends classical \prolog with constructs specific for defining probability distributions as well as with stratified negation. It uses the symbol \pneg for negation, where `$\backslash$' stands for `not' and `+' stands for `provable'. That is,  \pneg represents {\em negation as failure}. Stratification in logic programs and rule-based query languages is a well-known idea (see~\cite{AHV,AptBW88} and references there). For the convenience of the readers, let us recall its definition. We actually define a core sublanguage of \prolog with stratified negation, common to practically all logic programming languages, so will refer to it as  {\em stratified logic programs}.

A {\em stratified logic program} is a finite set of rules of the following form:\footnote{We restrict the definitions to the propositional case. Their extension to the 1st-order case is given in Section~\ref{sec:fo}. For details, see also, e.g.,~\cite{AHV}.}
\begin{equation}\label{eq:prologrule}
 h\when \pm r_1, \ldots, \pm r_k.
\end{equation}
where $k\geq 0$, and:
\begin{itemize}
    \item $\pm$ stands for the negation \pneg  or the empty symbol; occurrences of $r$ of the form $\pneg r$ are called {\em negative}, and when not preceded by \pneg, we call them {\em positive};
    \item $h, r_1,\ldots,r_k$ are propositional variables;
    $h$ is called the {\em head} or the {\em conclusion} of~\eqref{eq:prologrule}, $\pm r_1, \ldots, \pm r_k$ are called the {\em body} or {\em premises} of~\eqref{eq:prologrule}. 
\end{itemize} 
The following {\em stratification condition} is assumed, where by a {\em definition of a variable} $r$ we mean all of a program's rules with $r$ in the head:
\begin{quote}
    the set of propositional variables occurring in the program can be partitioned into a~finite collection of nonempty disjoint subsets $R_1, \ldots, R_n$ ($n\geq 1$), called {\em strata}, such that:
    \begin{itemize}
        \item  the definition of each variable $r$ occurring in the program is contained in one of the strata $R_1, \ldots, R_n$; 
        \item whenever a variable $r$ occurs (positively) as a premise of a program's rule with head in stratum $R_i$ then $r$ is to be defined in a stratum $R_{j}$ where $1\leq j\leq i$;
        \item whenever $\pneg r$ occurs as a premise of a  program's rule with head in stratum $R_i$ then $r$ is to be defined in a stratum $R_{j}$ where $1\leq j< i$.
    \end{itemize}
\end{quote}
That is, when a positive premise $r$ occurs in a rule, it has to be  defined in the current or an ``earlier'' stratum, and when a negative premise $\pneg r$ occurs in a rule, it has to be defined in  a~strictly ``earlier'' stratum. The essence of stratification is that the value of a propositional variable can be evaluated before being used in a negative premise of any rule.

The semantics of stratified logic programs is standard~\cite{AHV,AptBW88} (definable as the least wrt $\subseteq$ model of rules understood as implications, i.e., \eqref{eq:prologrule} being  understood as $(\pm r_1\land \ldots\land  \pm r_k)\implies h$).

The  \problog  constructs specific for defining probability distributions used in the paper are listed below. For simplicity, the presentation is restricted to propositional logic, since an extension to a 1st-order language over finite domains, which is defined later in the paper, is straightforward. For additional features, the reader is referred to related literature, including~\cite{deraedt-book,RaedtK15}, in addition to other references there.

\begin{itemize}
    \item To specify {\em probabilistic variables}, an operator `::' is used.\footnote{Probabilistic variables are also sometimes called {\em probabilistic facts}.} The specification $u::p$
    states that $p$ is true with probability $u$ and false with probability $(1-u)$.
    Probabilistic variables can be used to specify probabilistic facts and heads of probabilistic rules. 
    
    \item {\em Annotated disjunctions} support choices. An {\em annotated disjunction} is an expression of the form $u_1\!::\! p_1; \ldots ; u_k\!::\! p_k$, where $u_1,\ldots,u_k\in[0.0,1.0]$ such that $0.0\leq u_1+\ldots+u_k\leq 1.0$,  and $p_1,\ldots p_k$ are propositional variables.  If  $u_1+\ldots+u_k= 1.0$ then exactly one of $u_1\!::\! p_1, \ldots,$ $u_k\!::\! p_k$ is chosen as a~probabilistic fact. If $u_1+\ldots+u_k< 1.0$ then there is additionally a null choice with probability $1-(u_1+\ldots+u_k)$.
    
    \item {\em Queries} are used to compute probabilities. The probability of $p$ is returned as the result of the query $query(p)$ or $subquery(p,P)$, where the real typed variable $P$ obtains the probability of propositional variable $p$.
\end{itemize}

\problog\ {\em programs} used in this paper are stratified logic programs extended with a specification of a probability distribution expressed by a finite set of probabilistic variables, annotated disjunctions, and allowing queries over atomic formulas.\footnote{\problog contains  other constructs, too. Here we restricted its definition to constructs used in the paper.}

{\em Distribution semantics}~\cite{sato95}, used in \problog, assumes that probabilistic variables are independent, so probabilities of worlds are evaluated as the product of probabilities $u_i$ of probabilistic facts being true in these worlds and $(1-u_i)$ of those being false. More precisely, let $u_1::p_1,\ldots,u_m::p_m$ be all probabilistic facts specified directly or chosen from annotated disjunctions in the probability distribution specification part of a \problog program $\pi$. Let $\calw^\pi$ denote the set of all worlds assigning truth values to $p_1,\ldots,p_m$ (in addition to other propositional variables occurring in $\pi$). The probability structure {\em induced by} $\pi$ is $\calm^\pi=\tuple{\calw^\pi, 2^{\calw^\pi}, \calp^\pi}$, where probabilities of worlds are defined by:\footnote{Formally, probability $\calp^\pi$ is defined on events. Since each world $w$  can be considered as a singleton event $\{w\}$, for simplicity rather than  $\calp^\pi(\{w\})$, we write $\calp^\pi(w)$.}
\[\calp^\pi(w)\defeq\!\!\!\!\!\!\displaystyle\prod_{1\leq i \leq m, w\models p_i}\!\!\!\!\!\!\!\!u_i\;\;\;\; *\!\!\prod_{1\leq i \leq m, w\not\models p_i}\!\!\!\!\!\!(1-u_i),\]
and the probabilities of events, $e\subseteq\calw^\pi$, by $\displaystyle \calp^\pi(e)\defeq\sum_{w\in e}\calp^\pi(w).$

{\em Probabilities of queried propositional variables} are defined as the sum of probabilities of worlds, where they are true, 
\[\calp^\pi(p)=\calp^\pi(p^{\calm^\pi})\defeq\!\!\!\!\!\!\sum_{w\in\calw^\pi:\ w\models p}\!\!\!\!\!\!\calp^\pi(w).\]
Similarly, when $A$ is an arbitrary propositional formula, the {\em probability of} $A$ is, in distribution semantics, defined by:
\[\calp^\pi(A)\defeq\!\!\!\!\!\!\sum_{w\in\calw^\pi:\ w\models A}\!\!\!\!\!\!\calp^\pi(w).\]

In the rest of the paper, whenever we refer to $\calp^\pi$, the program $\pi$ is known from the context. To simplify notation, in such cases we omit the superscript $\pi$.

\begin{remark}\label{remark-compprop1st}
Observe that \problog does not support direct queries that allow one to compute probabilities of propositional or 1st-order formulas. To express such queries, transformations of propositional formulas to \problog rules are defined later in the paper, in Table~\ref{tab:transform} (page~\pageref{tab:transform}), and for 1st-order formulas -- in Table~\ref{tab:transformI} (page~\pageref{tab:transformI}).    \done
\end{remark}

Approximate evaluation of probabilities, with sampling, is available through \problog's interpreter accessible from:\\ \centerline{\small\url{https://dtai.cs.kuleuven.be/problog/index.html#download}}
using the command:\\ 
\centerline{\small \texttt{problog sample model.txt\;--estimate\;-N\;n}}
where `{\small\verb|model.txt|}' is a text file with a \problog source, and `{\small\verb|n|}' is a positive integer fixing the number of samples. For more information on inference by sampling in \problog see~\cite{RaedtK15}~and references there.

An online \problog interpreter sufficient to run programs included in this paper can be accessed via:\ \  {\small\url{https://dtai.cs.kuleuven.be/problog/editor.html}.}

\section{Model-Counting-Based Loss Measures of~Forgetting }\label{sec:modcountmeas}

\subsection{Introducing Model Counting-Based Measures} \label{sec:intromodcount}

 Recall that  $mod(\calt)$ denotes the set of models of \calt. Furthermore, let $\#\calt$ denote the number of models of $\calt$. Though the vocabulary after applying forgetting operators is a~subset of the vocabulary of the original theory \calt, for the sake of uniformity of presentation it is assumed that the considered worlds, thus models too, are built over the vocabulary of \calt. As an immediate corollary of~\eqref{eq:forgetimplies} one then has that: 
\begin{equation}\label{eq:inclusion}
mod\big(\sforget{Th(\bar{p},\bar{q});\bar{p}}\big)\!\subseteq\! mod\big(Th(\bar{p},\bar{q})\big)\!\subseteq\!  mod\big(\nforget{Th(\bar{p},\bar{q}); \bar{p}}\big),
\end{equation}
and so also:
\begin{equation}\label{eq:inclusion1}
    \#\sforget{Th(\bar{p},\bar{q});\bar{p}}\leq \#Th(\bar{p},\bar{q})\leq  \#\nforget{Th(\bar{p},\bar{q}); \bar{p}}.
\end{equation}

For any propositional theory $\calt$, the {\em probability of} \calt, $\prob{\calt}$, is defined in Definition~\ref{def:probtheory}. In the case of model counting-based measures we use uniform distribution on worlds, so the probability function is defined  by Equation~\eqref{eq:probth} (see also Remark~\ref{rem:probability}). 
Model counting-based measures of forgetting can now be defined as follows.

\begin{definition} \label{def:mcmeasures}
By {\em model counting-based loss measures of forgetting}, we understand:
    \begin{align}
	& \nlossm{\calt,\bar{p}}\defeq \prob{\nforget{\calt;\bar{p}}}-\prob{\calt}; \label{eq:nlossm}\\[-.4em]
	& \slossm{\calt,\bar{p}}\defeq \prob{\calt}-\prob{\sforget{\calt;\bar{p}}};  \label{eq:slossm} \\[-.4em]
	& \tlossm{\calt,\bar{p}}\defeq \prob{\nforget{\calt;\bar{p}}}-\prob{\sforget{\calt;\bar{p}}},\label{eq:tlossm}
\end{align}
where the underlying distribution on worlds is assumed to be uniform.
\done
\end{definition}
 
\noindent Obviously, 
\begin{equation}\label{eq:losst}
\tlossm{\calt,\bar{p}}=\nlossm{\calt;\bar{p}}+\slossm{\calt;\bar{p}}.
\end{equation}

\noindent The intuition behind the $loss_m$ measures is the following:
\begin{itemize}
    \item \eqref{eq:nlossm} indicates how much, in terms of the number of models, strong forgetting differs from 
     the original theory. In this case, the new theory is inferentially weaker than or equal to (wrt necessary conditions) the original theory, since certain necessary conditions that were previously entailed by the original theory might no longer be entailed by the resultant theory.
    \item  \eqref{eq:slossm} indicates how much, in terms of the number of models,    weak forgetting differs from  the original theory. In this case, the new theory is inferentially weaker than or equal to (wrt sufficient conditions) the original theory, since certain sufficient conditions that previously entailed the original theory might no longer entail the resultant theory;
    \item \eqref{eq:tlossm} indicates what the total loss is measured as the size of the gap between strong and weak forgetting.
\end{itemize}

\begin{remark}\label{rem:general}
    Note that Definition~\ref{def:mcmeasures} is very general. To work properly, it only uses the property expressed by Equation~\eqref{eq:important}. This practically holds for all logics where sufficient and necessary conditions can be defined. In such cases, sufficient conditions for \calt (thus also \sforget{\calt;.}) imply \calt which, in turn, implies necessary conditions of \calt (thus also \nforget{\calt;.}). \done
\end{remark}

To illustrate loss measures consider the following example.

\begin{example}\label{ex:prodline}
A production line is equipped, among others, with three highly specialized and expensive sensor platforms $sp_1, sp_2, sp_3$ controlling safety of the line equipment. To ensure  safety, it is required that that both $sp_1$ and  $sp_2$ together, or $sp_3$ itself, work correctly, sending alerts when an intervention/adjustment of production line parameters is needed. This condition is expressed by:
\begin{equation}\label{eq:propsensorp}
    \calt(sp_1,sp_2,sp_3)\defequiv (sp_1\land sp_2)\lor sp_3.
\end{equation} 
As a result of a bug in the control software causing an overvoltage,  $sp_1$ and $sp_3$ stopped functioning. Since replacing both of them right away is not possible with the current budget, the managers have to choose between an immediate replacement of either $sp_1$ or $sp_3$. In consequence, the remaining malfunctioning sensor platform is to be forgotten from the control belief base with the possibly smallest loss in reasoning capabilities (inferential strength). To support the decision making, loss functions can be used. By Theorem~\ref{thm:forget} and formulas~\eqref{eq:exists}, and \eqref{eq:forall} together, with obvious equivalence preserving transformations of propositional formulas, we have: 
\begin{itemize}
    \item $\nforget{\calt(sp_1,sp_2,sp_3); sp_1}  \equiv \exists sp_1\big((sp_1\land sp_2)\lor sp_3)\big)\equiv sp_2\lor sp_3$;
    \item $\sforget{\calt(sp_1,sp_2,sp_3); sp_1}  \,\equiv \forall sp_1\big((sp_1\land sp_2)\lor sp_3)\big)\equiv sp_3$;
    \item $\nforget{\calt(sp_1,sp_2,sp_3); sp_3}  \equiv \exists sp_3\big((sp_1\land sp_2)\lor sp_3)\big)\equiv \true$;
    \item $\sforget{\calt(sp_1,sp_2,sp_3); sp_3}  \,\equiv \forall sp_3\big((sp_1\land sp_2)\lor sp_3)\big)\equiv sp_1\land sp_2$.
\end{itemize}
Notice that, for $i\in\{1,3\}$, $\nforget{\calt(sp_1,sp_2,sp_3); sp_i}$ serves as a theory from which one can infer properties that are also inferred from \eqref{eq:propsensorp} but, of course, not all of them. For example, formula $sp_1\lor sp_3$ can be inferred from \eqref{eq:propsensorp} but not from any of $\nforget{\calt(sp_1,sp_2,sp_3); sp_i}$. The role of $\sforget{\calt(sp_1,sp_2,sp_3); sp_i}$ is different, as it is used as the weakest sufficient condition for \eqref{eq:propsensorp}. That is, to maintain the property \eqref{eq:propsensorp} after forgetting $sp_1$ or $sp_2$, one is interested in formulas implying $\sforget{\calt(sp_1,sp_2,sp_3); sp_i}$. Of course, $sp_3$ implies \eqref{eq:propsensorp}, but no longer implies $\sforget{\calt(sp_1,sp_2,sp_3); sp_i}$, so in this case the inferential capabilities decrease, too.

The probabilities needed to compute loss measures are: 
\[
\begin{array}{ll}
\prob{\calt(sp_1,sp_2,sp_3)}=0.625,\\
\prob{\nforget{\calt(sp_1,sp_2,sp_3); sp_1}}=0.75, &
\prob{\sforget{\calt(sp_1,sp_2,sp_3); sp_1}}=0.5, \\
\prob{\nforget{\calt(sp_1,sp_2,sp_3); sp_3}}=1.0, &
\prob{\sforget{\calt(sp_1,sp_2,sp_3); sp_3}}=0.25.
\end{array}
\]
Thus:
\[
\begin{array}{lclcl}
  \nlossm{\calt(sp_1,sp_2,sp_3),sp_1}&=&0.75-0.625&=&0.125, \\ 
  \slossm{\calt(sp_1,sp_2,sp_3),sp_1}&=&0.625-0.5&=&0.125,\\
  \tlossm{\calt(sp_1,sp_2,sp_3),sp_1}&=&0.75-0.5&=&0.25,\\
  \nlossm{\calt(sp_1,sp_2,sp_3),sp_3}&=&1.0-0.625&=&0.375, \\ 
  \slossm{\calt(sp_1,sp_2,sp_3),sp_3}&=&0.625-0.25&=&0.375,\\
  \tlossm{\calt(sp_1,sp_2,sp_3),sp_3}&=&1.0-0.25&=&0.75.
\end{array}
\]
The loss of inferential strength is greater (wrt to all three measures)  when $sp_3$ is forgotten than in the case of forgetting $sp_1$. From the point of view of preserving as much inferential strength as possible, one should then chose to replace $sp_3$ first and wait with $sp_1$ until the budget allows for its replacement.    \done
\end{example}

Proposition~\ref{prop:properties} below states some interesting properties of the defined loss measures, showing that they behave naturally wrt the intuitive properties of the forgetting operators. In particular, Point 2 shows that  nothing is lost when one forgets redundant variables (not occurring in the theory), thus also from \false or \true. Point~3 shows that  $\lossm{}$ is $0.0$ when nothing is forgotten; and Point~4 shows that the more propositional variables are forgotten, the greater $\lossm{}$ may~be.

\begin{proposition}\label{prop:properties}
 Let $*\in\{NC, SC, T\}$. Then for every propositional theory $\calt$,
   \begin{enumerate}
       \item $\lossm{\calt,\bar{p}}\in[0.0, 1.0]$;
       \item $\lossm{\calt,\bar{p}}=0.0$, when variables from $\bar{p}$ do not occur in $\calt$; in particular:\\[-1.5em]
       \begin{enumerate}
        \item $\lossm{\false,\bar{p}}=0.0$;
        \item $\lossm{\true,\bar{p}}=0.0$;      
       \end{enumerate}
       \item $\lossm{\calt,\bar{\epsilon}}=0.0$, where $\bar{\epsilon}$ stands for the empty tuple;
       \item $\lossm{\calt,\bar{p}}\leq \lossm{\calt,\bar{q}}$ when all propositional variables occurring in $\bar{p}$ also occur in $\bar{q}$ (\lossm{} is monotonic wrt forgotten vocabulary).
   \end{enumerate}
\end{proposition}

\noindent \textbf{Proof}:
\begin{itemize}
	\item Point 1 directly follows from Definition~\ref{def:mcmeasures} and Equation~\eqref{eq:important}.
	\item Points 2 and 3 are consequences of characterizations given in Theorem~\ref{thm:forget}. Indeed, when variables from $\bar{p}$ do not occur in $\calt$, in particular when $\bar{p}=\bar{\epsilon}$, both $\exists\bar{p}$ and $\forall\bar{p}$ occurring at the righthand sides of Points 1 and 2 of Theorem~\ref{thm:forget} become redundant, thus $\nforget{\calt; \bar{p}}$ and $\sforget{\calt; \bar{p}}$ are equivalent to $\calt$, so $\prob{\nforget{\calt;\bar{p}}}=\prob{\sforget{\calt;\bar{p}}}=\prob{\calt}$.
	\item Point 4 also follows  from Theorem~\ref{thm:forget}, since for variables of  $\bar{p}$ included in variables of $\bar{q}$,\\[-1.5em]
	\begin{itemize}
		\item $\models \exists\bar{p}\,\calt\implies\exists\bar{q}\, \calt $, so 
		$\models \nforget{\calt;\bar{p}}\implies\nforget{\calt;\bar{q}}$.  Therefore,\\ $\prob{\nforget{\calt;\bar{p}}}\leq \prob{\nforget{\calt;\bar{q}}}$, so $\nlossm{\calt,\bar{p}}\leq\nlossm{\calt,\bar{q}}$; 
		\item $\models \forall\bar{q}\,\calt\implies\forall\bar{p}\, \calt $, so 
		$\models \sforget{\calt;\bar{q}}\implies\sforget{\calt;\bar{p}}$.  Therefore,\\ $\prob{\sforget{\calt;\bar{p}}}\geq \prob{\sforget{\calt;\bar{q}}}$, so $\slossm{\calt,\bar{p}}\leq\slossm{\calt,\bar{q}}$;
		\item the fact that $\tlossm{\calt,\bar{p}}\leq\tlossm{\calt,\bar{q}}$ follows directly from the above using the property~\eqref{eq:losst}. \done
	\end{itemize} 
\end{itemize} 

 For \nlossm{} the property of {\em countable additivity} holds, and is formulated in the following proposition.\footnote{Countable additivity, together with nonnegativity (following from Point 1), and Point 2(a) of Proposition~\ref{prop:properties}, are traditional axioms of measures -- see, e.g.,~\cite[Chapter 2]{Billingsley}.} 

\begin{proposition} \label{prop:propnc}
Assume that for all $1\!\leq\! i\!\not=\!j$, $\calt_i$ and $\calt_j$ mutually exclude one another, i.e., $\calt_i\land \calt_j\equiv\false$. Then:
 $\displaystyle \nlossm{\bigvee_{1\leq i}\!\calt_i,\bar{p}}=\sum_{1\leq i}\!\nlossm{\calt_i,\bar{p}}$.
\end{proposition}

\noindent \textbf{Proof}:
The proof can be carried out as follows:
\[
\begin{array}{lcll}
	\displaystyle \nlossm{\bigvee_{1\leq i}\!\calt_i,\bar{p}}&\!\!\! =\!\!\! & \displaystyle \prob{\nforget{\bigvee_{1\leq i}\!\calt_i;\bar{p}}}-\prob{\bigvee_{1\leq i}\!\calt_i}= &  \mbox{(by Theorem~\ref{thm:forget}.1)}\\
	& = & \displaystyle \prob{\exists\bar{p}\bigvee_{1\leq i}\,\calt_i}-\prob{\bigvee_{1\leq i}\!\calt_i}= & \hspace*{-1.1cm}\displaystyle \prob{\bigvee_{1\leq i}\exists\bar{p}\,\calt_i}-\prob{\bigvee_{1\leq i}\!\calt_i}=\\
	& = & \displaystyle \prob{\bigvee_{1\leq i}\nforget{\calt_i;\bar{p}}}-\prob{\bigvee_{1\leq i}\!\calt_i}.
\end{array}
\]  
Since probability is a measure, we have $\displaystyle \prob{\bigvee_{1\leq i}\nforget{\calt_i;\bar{p}}}=\sum_{1\leq i}\prob{\nforget{\calt_i;\bar{p}}}$ and \break $\displaystyle \prob{\bigvee_{1\leq i}\!\calt_i}=\sum_{1\leq i}\prob{\calt_i}$. Therefore:
 \[
\begin{array}{lcll}
 \displaystyle \prob{\bigvee_{1\leq i}\nforget{\calt_i;\bar{p}}}-\prob{\bigvee_{1\leq i}\!\calt_i}& \!\!\!= \!\!\!& \displaystyle \sum_{1\leq i}\prob{\nforget{\calt_i;\bar{p}}}- \sum_{1\leq i}\prob{\calt_i}=\\
	&\!\!\!= \!\!\!& \displaystyle \sum_{1\leq i}\big(\prob{\nforget{\calt_i;\bar{p}}}-\prob{\calt_i}\big)= & \!\!\!\!\!\!\!\displaystyle\sum_{1\leq i}\nlossm{\calt_i,\bar{p}}.\ \done
\end{array}
\]

For the measures \slossm{} and \tlossm{} countable additivity does not hold in general. However, a weaker property, {\em countable subadditivity} does hold, as formulated in the next proposition.

\begin{proposition}\label{prop:subadd}
Assume that for all $1\!\leq\! i\!\not=\!j$, $\calt_i$ and $\calt_j$ mutually exclude one another, i.e., $\calt_i\land \calt_j\equiv\false$. Then, for $*\in\{SC, T\}$,
 $\displaystyle \lossm{\bigvee_{1\leq i}\!\calt_i,\bar{p}}\leq\sum_{1\leq i}\!\lossm{\calt_i,\bar{p}}$.
\end{proposition}
\noindent \textbf{Proof}:
\begin{itemize}
    \item Let us start with the case when $*=SC$. Notice that $\displaystyle \bigvee_{1\leq i}\big(\forall\bar{p}\calt_i\big)$ implies $\displaystyle\forall\bar{p}\big(\bigvee_{1\leq i}\calt_i\big)$.
Thus, $\displaystyle\prob{\bigvee_{1\leq i}\big(\forall\bar{p}\calt_i\big)}\leq\prob{\forall\bar{p}\big(\bigvee_{1\leq i}\calt_i\big)}$. Therefore, 
\[
\prob{\bigvee_{1\leq i}\calt_i}-\prob{\bigvee_{1\leq i}\underbrace{\big(\forall\bar{p}\calt_i\big)}_{\displaystyle\sforget{\calt_i;\bar{p}}}}\geq 
\underbrace{\prob{\bigvee_{1\leq i}\calt_i}-\prob{\underbrace{\forall\bar{p}\big(\bigvee_{1\leq i}\calt_i\big)}_{\displaystyle\sforget{\bigvee_{1\leq i}\calt_i;\bar{p}}}}}_{\displaystyle\slossm{\bigvee_{1\leq i}\calt_i,\bar{p}}},
\]
so, using the assumption as to the mutual exclusion of $\calt_i$ and $\calt_j$ ($i\not=j$) and countable additivity of probability, we have
$\displaystyle
\sum_{1\leq i}\prob{\calt_i}-\sum_{1\leq i}\prob{\sforget{\calt_i;\bar{p}}}\geq \slossm{\bigvee_{1\leq i}\calt_i,\bar{p}}
$. One then concludes that  $\displaystyle\sum_{1\leq i}\underbrace{\big(\prob{\calt_i}-\prob{\sforget{\calt_i;\bar{p}}}\big)}_{\displaystyle\slossm{\calt_i,\bar{p}}}\geq \slossm{\bigvee_{1\leq i}\calt_i,\bar{p}}$, what was to be shown. 
\item 
The case when $*=T$ is now obvious, using~\eqref{eq:losst} together with Proposition~\ref{prop:propnc} and the just proven countable subadditivity of \slossm{}.\done
\end{itemize}

\subsection{Computing Model Counting-Based Loss Measures Using \problog}\label{sec:compmodcount}

To compute the probability of \calt, \problog is used. Though many \#SAT solvers counting the number of satisfying assignments exist -- see~\cite{ChakrabortyMV21,fichte,GomesSS21,SoosM19} and references there, \problog has been chosen, since:
\begin{itemize}
    \item it allows for computing exact probability as well as probability based on approximate (sampling based) evaluation;
    \item the resulting probability and loss measures can be used within a probabilistic logic programming framework in a~uniform manner. 
\end{itemize}

To illustrate the idea of computing probabilities of propositional formulas, not yet introducing formal transformations into \problog programs, let us start with the following example example.

\begin{program}[ht]
  \caption{\problog program for computing  the~probability of $\calt_c$.\label{prog:tcar} }

  $0.5::car.$\hspace*{3mm} $0.5::reliable.$\ \  $0.5::fast.$\;
  $0.5::fcar.$\hspace*{1.25mm} $0.5::jcar.$ \hspace*{4.6mm}    $0.5::ecar.$\;
  $r \when\; car, reliable, fcar.$ \label{line:a}\;
  $r \when\; \pneg jcar.$  \hspace*{20mm} \% $r$ represents \eqref{eq:cars1}\label{line:aa}\;
  $s \when\; car, fast, fcar.$ \;
  $s \when\; \pneg ecar.$  \hspace*{20mm} \% $s$ represents \eqref{eq:cars2}\;
  $u \when\; jcar.$\;
  $u \when\; ecar.$\;
  $u \when\; \pneg fcar.$ \hspace*{19mm} \% $u$ represents \eqref{eq:cars3}\;
  $t \when\; r, s, u.$    \hspace*{22mm} \% $t$ represents \eqref{eq:cars1}$\,\land\,$\eqref{eq:cars2}$\,\land\,$\eqref{eq:cars3}\label{line:e}\;
\end{program}

\begin{example}\label{ex:cars-mod-count}
The theory  $\calt_c$ expressed by formulas~\eqref{eq:cars1}--\eqref{eq:cars3} is encoded as shown in Program~\ref{prog:tcar} computing the probability of this theory. 
Indeed, since  we deal with complete states, negation \pneg is equivalent to $\lneg$, and rules \ref{line:a}--\ref{line:aa} represent the disjunction $\lneg jcar \lor (car\land  reliable \land  fcar)$, we have that $r$ is equivalent to~\eqref{eq:cars1}. Similarly, the variable 
$s$ is equivalent to
$\lneg ecar\lor ( car\land fast\land fcar)$, i.e., to~\eqref{eq:cars2},
and the variable $u$ is equivalent to
$\lneg fcar\lor jcar\lor ecar$, i.e., to~\eqref{eq:cars3}. Since $t$ is the conjunction of $r$, $s$ and $u$, it is equivalent to $\calt_c$. Therefore, the probability of $t$, computed by \problog's interpreter, is equal to $\prob{\calt_c}$. Indeed, there are $6$ propositional variables and each possible world encodes an assignment of truth values to these propositions with the probability $0.5$. So there are $2^6$ possible worlds, each having the probability $\frac{1}{2^6}$, and $t$ is true in $13$ of them (in those satisfying $\calt_c$). Accordingly, given $query(t)$, the \problog interpreter returns the answer that the probability of $t$ (i.e., of $\calt_c$), is $0.203125$.\done 
\end{example}

The following shows how this approach works in general. 

First, one has to define a transformation of an arbitrary propositional formula into a set of rules. Let $A$ be a propositional formula. The rules of a {\em program $\pi^0_A$ encoding} $A$, are constructed as shown in Table~\ref{tab:transform}, where the $r$ indexed with formulas are auxiliary propositional variables. The intended meaning of each auxiliary variable $r_B$ is that it is equivalent to  formula $B$. It is assumed that whenever a subformula occurs more than once, a single auxiliary variable is used to substitute all of its occurrences. 

\begin{table}[ht]
\caption{Transformation of an arbitrary propositional formula to a~stratified logic program.\label{tab:transform}}
\centering
\begin{tabular}{|c|}\hline\mbox{}\\[-0.5em]
\qquad $
\!\!\pi^0_A\! \defeq \left\{\begin{array}{ll}
   \!\!\!\!\begin{array}{l}\emptyset \mbox{\ (the empty program);}\\[-0.8mm]
   r_A\!\defeq\! A\end{array} & \hspace*{-3mm}\mbox{when } A \mbox{ is a propositional variable};\\[-1.5mm] 
   \!- - - - - \\[-1.5mm] 
   \!\!\!\!\begin{array}{l}
        \pi^0_B. \\[-0.8mm]
        r_A\when \pneg r_B.  
   \end{array}
      & \hspace*{-3mm} \mbox{when } A=\lneg B;\\[-1.5mm]  
   \!- - - - - \\[-1.5mm] 
   \!\!\!\!\begin{array}{l}
   \pi^0_C.\; \pi^0_D. \\[-0.8mm]
   r_A\when r_C, r_D.\end{array} 
   & \hspace*{-3mm} \mbox{when } A=C\land D;\\[-1.5mm] 
   \!- - - - - \\[-1.5mm] 
   \!\!\!\!\begin{array}{l}\pi^0_C.\; \pi^0_D. \\[-0.8mm]
   r_A\when r_C.\\[-0.8mm]
   r_A\when r_D.\end{array} & \hspace*{-3mm} \mbox{when } A=C\lor D;\\[-1.5mm] 
   \!- - - - - \\[-1.5mm] 
   \!\!\!\!\begin{array}{l}\pi^0_C.\; \pi^0_D. \\[-0.8mm]
   r_A\when \pneg r_C.\\[-0.8mm]
   r_A\when r_D.\end{array} & \hspace*{-3mm} \mbox{when } A=C\implies D;\\[-1.5mm] 
   \!- - - - - \\[-1.5mm] 
   \!\!\!\!\begin{array}{l}\pi^0_C.\; \pi^0_D. \\[-0.8mm]
   r_{A'}\when \pneg r_C.\\[-0.8mm]
   r_{A'}\when r_D.\\[-0.8mm]
   r_{A''}\when \pneg r_D.\\[-0.8mm]
   r_{A''}\when r_C.\\[-0.8mm]   
   r_A\when r_{A'}, r_{A''}.
   \end{array} & \hspace*{-3mm} \mbox{when } A=C\equiv D.
\end{array}\right.
$ 
\\ \qquad\quad\\
\hline
\end{tabular}
\end{table}

The following example illustrates the transformation defined in Table~\ref{tab:transform}.

\begin{example}\label{ex:table-transform-prop}
   As an example of a transformation defined in Table~\ref{tab:transform}, consider Program~\ref{prog:example} transforming formula $\big(q\equiv (p\land\lneg q\land s)\big)$ into a set of rules. Note that the number of auxiliary variables are often optimized while encoding longer expressions consisting of the same commutative connectives, such as a~conjunction with more than two arguments, $A_1\land\ldots\land A_k$, with $k>2$. In this case, one usually uses a single variable together with the rule $r_{A_1\land\ldots\land A_k}\when r_{A_1},\ldots,r_{A_k}$. An example of such an optimization is demonstrated in  Line~\ref{line:exampleand} of Program~\ref{prog:example} which contains a~conjunction of three literals. Similarly, one optimizes auxiliary variables in disjunctions with more than two arguments. \done 
\end{example}

\begin{program}[t]
  \caption{A program $\pi^0_{q\equiv (p\land\lneg q\land s)}$ encoding the formula $\big(q\equiv (p\land\lneg q\land s)\big)$.\label{prog:example} }
 \% $\pi^0_p$, $\pi^0_q$, $\pi^0_s$ are empty; $r_p\!\defeq\! p$, $r_q\!\defeq\! q$, $r_s\!\defeq\! s$\;
 $r_{\lneg q}\when\;\pneg q.$\;
 $r_{p\land\lneg q\land s}\when\; p, r_{\lneg q}, s.$\label{line:exampleand}\;
 $r_{(q\equiv (p\land\lneg q\land s))'} \when\; \pneg q.$ \;
 $r_{(q\equiv (p\land\lneg q\land s))'} \when\; r_{p\land\lneg q\land s}.$\;
 $r_{(q\equiv (p\land\lneg q\land s))''}\when\; \pneg r_{p\land\lneg q\land s}.$\;
 $r_{(q\equiv (p\land\lneg q\land s))''}\when\; q.$ \label{line:rw2}\;
 $r_{q\equiv (p\land\lneg q\land s)}\when\; r_{(q\equiv (p\land\lneg q\land s))'}, r_{(q\equiv (p\land\lneg q\land s))''}.$
\end{program}

We have the following proposition.

\begin{proposition}\label{prop:linear}
    For every propositional formula $A$, the size of $\pi^0_A$ is linear in the size of $A$.
\end{proposition}
\noindent \textbf{Proof}: By construction of $\pi^0_A$, 
for each connective occurring in $A$ at most $5$ additional rules are added. If there are $n$ connectives then $\pi^0_A$ contains at most $5n$ rules.
\done

Similarly, it is easily seen that the following proposition also holds.

\begin{proposition}\label{prop:determines}\mbox{}
\begin{enumerate}
    \item Let  $A$ be a  propositional formula. Then each world $w\in\calw_A$ determines the values of auxiliary variables in $\pi^0_A$. 
    \item Given a world $w\in\calw_A$, the truth values of auxiliary variables in $\pi^0_A$ can be determined in time linear in the size of $A$.\done
\end{enumerate}
\end{proposition}

The following lemmas are used to show Theorem~\ref{thm:probA}.

\begin{lemma}\label{lemma:strat}
   For every propositional formula $A$, $\pi^0_A$ is a~stratified logic program.
\end{lemma}
\noindent \textbf{Proof}: By construction of $\pi^0_A$: whenever a variable, say $r_F$, is negated by \pneg, it appears in a body of a rule with variable $r_E$ in its head where $F$ is a subformula of $E$. This provides a natural stratification reflecting the subformula nesting.\done

One can now define a {\em \problog program, $\Pi^m_A$, computing the probability of a formula $A$}. $\Pi^m_A$ is specified by the following probabilistic facts and rules:
\begin{equation}\label{eq:PiAm}
  \Pi_A^m\defeq\left|  \begin{array}{l}
   0.5::p_1.\; \ldots\; 0.5::p_k.   \\ 
   \pi^0_A.
\end{array}\right.
\end{equation}
In this case,   $p_1,\ldots, p_k$ are all  variables occurring in $A$ and the probability of $A$ is given by querying for the probability of the auxiliary variable $r_A$ representing the formula $A$ in the transformation defined in Table~\ref{tab:transform}. The variable $r_A$ is included in $\pi^0_A$, the stratified logic program associated with~$A$.

Note that:
\begin{itemize}
    \item $\Pi^m_A$ as specified by~\eqref{eq:PiAm} is a \problog program since, by Lemma~\ref{lemma:strat}, $\pi^0_A$ is a stratified logic program;
    \item the probabilities $0.5$, assigned to propositional variables together with artificially assuming them independent, are used as a technical fix that ensures a uniform distribution on worlds and that guarantees the program specifies the number of models where $A$ is true. 
\end{itemize}

\begin{lemma}\label{lemma:equiv}
    Let $\pi^0_A$ occur in the program $\Pi_A^m$. Then for every subformula $B$ of $A$ occurring in the construction of $\pi^0_A$, we have that $r_B$ is equivalent to $B$.
\end{lemma}
\noindent \textbf{Proof} 
By Lemma~\ref{lemma:strat}, $\pi^0_A$ specifies a stratified logic program. The semantics of stratified logic programs can be defined in terms of least models of rules expressed by implications~\cite{AptBW88}. In a~manner similar  to the  well-known completion technique~\cite{Clark77} (for its extension to stratified logic programs see, e.g.,~\cite{AptBW88}),
\begin{equation}\label{eq:observ}
    \parbox{13.5cm}{each $r_B$ appearing in $\pi_A^m$ in a head of a rule (or more rules, as in the case of $\lor, \implies$ and $\equiv$), is actually equivalent to the rule's body (respectively, disjunction of bodies) due to the completion.  This observation is used below when claiming equivalences, rather than implications directly reflecting rules.}
\end{equation}

\noindent One then proceeds by structural induction on $B$.\vspace*{-0.5em}

\hspace*{-0.9cm}\begin{itemize}
    \item  When $B$ is a propositional variable, say $p$, then $\pi^0_B$, being $\pi^0_p$, is empty, and $r_B\defeq p$, so the induction hypothesis is trivially true.
    \item When $B$ is a negated formula, say $\lneg E$, then $r_B$, being $r_{\lneg E}$, is defined by $r_{\lneg E}\when \pneg r_E$, where $\pneg$ stands for negation as failure. By observation~\eqref{eq:observ}, $r_{\lneg E}\equiv \pneg r_E$. Note that probabilistic clauses in $\Pi_A^m$, given in the first line of \eqref{eq:PiAm}, assign truth values to all propositional variables occurring in $A$. Therefore, in each world, the truth values of all subformulas of $A$ are uniquely determined, in which case the only reason for a failure to prove an $A$'s subformula, say $E$, is the falsity of $E$. That is, in $\Pi_A^m$, negation as failure becomes classical negation. Thus $r_B=r_{\lneg E}\equiv\pneg r_E\equiv\lneg r_E$. By the inductive assumption, $r_E$ is equivalent to $E$, so  $\lneg r_E$ is equivalent to $\lneg E$, what shows the equivalence of $r_B$ and~$B$.
    \item  When $B$ is a non-negated subformula of $A$ occurring in the construction of $\pi^0_A$, one uses the inductive assumption together with observation~\eqref{eq:observ} and obvious tautologies of propositional logic to show that the auxiliary variable $r_B$ is equivalent to $B$.  
\done 
\end{itemize}

The following theorem can now be shown, stating that programs $\Pi_A^m$ can be used to calculate the probability measures for a formula (theory) which are then used as a basis for computing the values of the considered model counting-based measures.

\begin{theorem}\label{thm:probA}
   For every propositional formula $A$, the probability of $A$ (in the sense of definition given by~\eqref{eq:probth}) is equal to the probability of $r_A$ given by the \problog program $\Pi^m_A$, defined by~\eqref{eq:PiAm}. 
\end{theorem}
\noindent \textbf{Proof}: By Lemma~\ref{lemma:equiv}, 
$r_A$ is equivalent to $A$, so in every possible world, $r_A$ is true iff $A$ is true. Assume $p_1,\ldots, p_k$ are all propositional variables occurring in $A$. There are $2^k$ possible worlds, so if $A$ is true in $n$ worlds, its probability is $\displaystyle\frac{n}{2^k}$. Each propositional variable is assigned the probability $0.5$ in $\Pi^m_A$, so the probability of each possible world  generated by $\Pi^m_A$ 
is $\displaystyle\frac{1}{2^k}$. Since $r_A$ is true exactly in worlds where $A$ is true and, by the distribution semantics, the probability of $r_A$  is the sum of probabilities of possible worlds where $r_A$ is true, the value of $r_A$ is:
\[r_A=\sum_{w\in\calw: w\models r_A}\frac{1}{2^k}=\frac{n}{2^k}=\prob{A}.\]

\vspace*{-2.5em}
\done
\mbox{}\\[0.3em]

In order to compute measures 
$\nlossm{\calt,\bar{p}}$, $\slossm{\calt,\bar{p}}$ and $\tlossm{\calt,\bar{p}}$, it now suffices to run programs $\Pi^m_{\nforget{\calt;\bar{p}}}$, $\Pi^m_{\sforget{\calt;\bar{p}}}$ and $\Pi^m_{\calt}$. The probabilities \prob{\nforget{\calt,\bar{p}}}, \prob{\sforget{\calt;\bar{p}}} and \prob{\calt} required in Definition~\ref{def:mcmeasures} are given respectively as the probabilities of $r_{\nforget{\calt;\bar{p}}}$, $r_{\sforget{\calt;\bar{p}}}$ and $r_{\calt}$.

\section{Probabilistic-based Loss Measures of~Forgetting}\label{sec:probmeas}

\subsection{Introducing Probabilistic-Based Measures}

Model counting-based loss measures are defined using a~uniform probability distribution on worlds. Probabilistic-based loss measures generalize this idea to allow arbitrary probability distributions on worlds. The idea is to reflect a type of reasoning that emphasizes more likely rather than less likely/unlikely worlds. In general, it is then assumed that an arbitrary probability distribution over worlds is provided, and one is  interested in probabilities of formulas wrt this distribution. 

Let us then assume that a probability distribution on worlds (understood as singleton events) is given, 
$
\calp_\calw: \calw\longrightarrow [0.0, 1.0],
$
and let \calt be a propositional theory.  The {\em probability of \calt wrt} $\calp_\calw$, $\calp_0(\calt)$, is then defined by:
\begin{equation}\label{eq:probtheory}
    \probz{\calt}\defeq\!\! \sum_{w\in\calw,\; w\models\calt}\!\!\!\!\probw{w}.
\end{equation}

Probabilistic-based loss measures of forgetting are defined as follows.

\begin{definition} \label{def:probmeasures}
By {\em probabilistic-based loss measures of forgetting wrt probability distribution} $\probz{}$, it is understood that:\\[-1.2em]
    \begin{align}
	& \nlossp{\calt,\bar{p}}\defeq \probz{\nforget{\calt;\bar{p}}}-\probz{\calt}; \label{eq:nlossp}\\[-0.4em]
	& \slossp{\calt,\bar{p}}\defeq \probz{\calt}-\probz{\sforget{\calt;\bar{p}}};  \label{eq:slossp} \\[-0.4em]
	& \tlossp{\calt,\bar{p}}\defeq \probz{\nforget{\calt\!;\bar{p}}}\!-\!\probz{\sforget{\calt\!;\bar{p}}},\label{eq:tlossp}
\end{align}
where the underlying distribution on worlds is assumed to be arbitrary.
\done
\end{definition}

As in the case of model counting-based loss measures, \[\tlossp{\calt,\bar{p}}=\nlossp{\calt,\bar{p}}+\slossp{\calt,\bar{p}}.\] 

It is worth emphasizing that, as indicated in Remark~\ref{rem:general}, the Definition~\ref{def:probmeasures} is very general, only using the property expressed by Equation~\eqref{eq:important}.

Notice that probability measures on worlds allow one to  weight the inferential strength according to probabilities of the involved propositional variables, which might be useful in decision making, as illustrated in the following example.

\begin{example}[Example~\ref{ex:prodline} continued] \label{ex:prodlinecont}
Assume that it takes time for the company in question to  adapt and approve the budget to cover the unexpected maintenance expenses. Assume further that the most intensive reasoning processes both about maintaining 
(in terms of sufficient conditions) and consequences of (in terms of necessary conditions) the control belief base happen when sensor platforms send alerts indicating the need of intervention/adjusting production line control parameters. When the inferential strength is considered, one is then interested in relative frequencies of the reasoning during the time period when the budget is being processed. Let, throughout that time period, the frequencies be expressed by probabilities with:
\begin{itemize}
    \item $sp_1$ being expected to send an alert with the probability $0.2$;
    \item $sp_2$ being expected to send an alert with the probability $0.3$;
    \item $sp_3$ being expected to send an alert with the probability $0.01$.
\end{itemize}
In this case the probabilities needed to compute loss measures are: 
\[
\begin{array}{ll}
\probz{\calt(sp_1,sp_2,sp_3)}= 0.0694,\\
\probz{\nforget{\calt(sp_1,sp_2,sp_3); sp_1}}=0.307, &
\probz{\sforget{\calt(sp_1,sp_2,sp_3); sp_1}}=0.01, \\
\probz{\nforget{\calt(sp_1,sp_2,sp_3); sp_3}}=1.0, &
\probz{\sforget{\calt(sp_1,sp_2,sp_3); sp_3}}=0.06.
\end{array}
\]
Thus:
\[
\begin{array}{lclcl}
  \nlossp{\calt(sp_1,sp_2,sp_3),sp_1}&=&0.307-0.0694&=&0.2376, \\ 
  \slossp{\calt(sp_1,sp_2,sp_3),sp_1}&=&0.0694-0.01&=&0.0594,\\
  \tlossp{\calt(sp_1,sp_2,sp_3),sp_1}&=&0.307-0.01&=&0.297,\\
  \nlossp{\calt(sp_1,sp_2,sp_3),sp_3}&=&1.0-0.0694&=&0.9306, \\ 
  \slossp{\calt(sp_1,sp_2,sp_3),sp_3}&=&0.0694-0.06&=&0.0094,\\
  \tlossp{\calt(sp_1,sp_2,sp_3),sp_3}&=&1.0-0.06&=&0.94.
\end{array}
\]
The loss of inferential strength wrt necessary conditions as well as the total loss when forgetting $sp_3$ is still greater than the loss when forgetting $sp_1$. However, the loss wrt sufficient  conditions is now greater when forgetting $sp_1$ than in the case of forgetting $sp_3$. The production managers then have to decide what is more important: the inference by sufficient conditions or maybe from the necessary ones. In many cases, for maintaining properties like~\eqref{eq:propsensorp}, it is more important than inferring their conclusions, since typically one is  more interested in conditions implying safety than in conditions that safety implies. In such circumstances the choice to replace $sp_1$ first appears to be more reasonable.
  \done
\end{example}

We have the following proposition analogous to the case of model counting-based measures. 

\begin{proposition}
For every propositional theory $\calt$, 
\begin{itemize}
    \item for $*\!\in\!\{NC, SC, T\}$, the measures \lossp{\calt,\bar{p}} enjoy properties analogous to those stated in Points 1--4 of Proposition~\ref{prop:properties}, where $loss^*_m$ is substituted by $loss^*_p$;
    \item the measure \nlossp{} satisfies countable additivity as expressed in Proposition~\ref{prop:propnc} with \nlossm{} substituted by \nlossp{}.
    \item for $*\in\{SC, T\}$, the measures \lossp{\calt,\bar{p}} satisfy countable subadditivity as expressed in Proposition~\ref{prop:subadd} with \lossm{}, \slossm{} and \tlossm{} substituted by \lossp{}, \slossp{} and \tlossp{}, respectively. \done
\end{itemize}
\end{proposition}

\subsection{Computing Probabilistic-Based Loss Measures Using \problog}\label{sec:compprob}

As shown through research with probabilistic logic programming and applications of distribution semantics, the probability $\calp_\calw$ can, in many cases, be specified by using \problog probabilistic facts and rules (see, e.g., \cite{deraedt-book,Pfeffer2016,riguzzi,SatoK97} and references there). 

Assuming that $\calp_\calw$ can be specified in \problog,   a~{\em \problog program, $\Pi^p_A$}, can be defined that computes the probability of $A$ wrt probability distribution $\calp_\calw$:
\begin{equation}\label{eq:PiAp}
  \Pi_A^p\defeq\left|  \begin{array}{l}
    \mbox{\problog\ specification of } \calp_\calw
     \mbox{ over  $p_1,\ldots, p_k$}.\\
    0.5::q_1.\; \ldots\; 0.5::q_n.   \\ 
   \pi^0_A.
\end{array}\right.
\end{equation}
In this case,   $p_1,\ldots, p_k$ are the pairwise independent probabilistic variables used to define the probability distribution $\calp_\calw$ over worlds, and $q_1,\ldots,q_n$ are the variables occurring in $A$  other than $p_1,\ldots, p_k$. The  probability of $A$ is given by querying for the probability of $r_A$ which is included in $\pi^0_A$, the stratified logic program associated with $A$.

Computing probabilistic-based loss measures 
$\nlossp{\calt,\bar{p}}$, $\slossp{\calt,\bar{p}}$ and $\tlossm{\calt,\bar{p}}$, can now be done by analogy with the method described in the last paragraph of Section~\ref{sec:compmodcount}, i.e., by running programs 
$\Pi^p_{\nforget{\calt;\bar{p}}}$, $\Pi^p_{\sforget{\calt;\bar{p}}}$ and $\Pi^p_{\calt}$. The probabilities \prob{\nforget{\calt,\bar{p}}}, \prob{\sforget{\calt;\bar{p}}} and \prob{\calt} required in Definition~\ref{def:probmeasures} are given respectively as the probabilities of $r_{\nforget{\calt;\bar{p}}}$, $r_{\sforget{\calt;\bar{p}}}$ and $r_{\calt}$. 

\begin{example}[Example~\ref{ex:cars-mod-count} continued]\label{ex-mod-count-cont-first}
As an example, consider the theory $\calt_c$ from Section~\ref{sec:intro}, and assume that in a district of interest, the probability of $ecar$ is $0.2$ and the probability of $jcar$ is $0.3$. Assume further that the choices of $jcar$ and $ecar$ exclude each other. So, in terms of definition~\eqref{eq:PiAp} $ecar$ plays the role of $p_1$ and $jcar$ plays the role of $p_2$, and there are no other $p_i$'s for $p\not\in\{1,2\}$. The other variables occurring in the theory $\calt_c$ play the role of $q$'s form the second line of program in~\eqref{eq:PiAp}. 
Program~\ref{prog:tcarprob} can now be used, where an annotated disjunction (Line~\ref{line:one}) is used to ensure a suitable probability distribution together with the mutual exclusion of $ecar, jcar$. In this case $\probz{\calt_c}$ is $0.3125$. \done
\end{example}

 \begin{program}[ht]
  \caption{\problog program for computing  the~probability of $\calt_c$\\ 
  \hspace*{1.85cm}when $\prob{ecar}=0.2$ and $\prob{jcar}=0.3$.\label{prog:tcarprob} }
  $0.2::ecar;\ \  0.3::jcar$.\ \ \ \ \ \ \  \% (an annotated disjunction)\label{line:one}\label{line:31}\;
  $0.5::car.$\hspace*{3mm} $0.5::reliable.$\ \  $0.5::fast.$\ \ $0.5::fcar.$\label{line:32}\;
  Lines \ref{line:a}--\ref{line:e} of Program~\ref{prog:tcar}
\end{program}

In summary, \problog is used to specify a given probability on worlds and then rules of \problog programs $\Pi^p_{\nforget{\calt;\bar{p}}}$, $\Pi^p_{\sforget{\calt;\bar{p}}}$ and $\Pi^p_{\calt}$ are used for computing probabilistic-based loss measures.

Notice that, as in Section~\ref{sec:compmodcount}, the probabilistic facts $0.5::q_i$ ($1\leq i\leq n$) serve as  technical fix that enables generation of all possible worlds, where propositional variables whose probabilities are not given explicitly are also included (Line 2 in program \eqref{eq:PiAp}, exemplified by Line 2 in Program~\ref{prog:tcarprob}). The following proposition shows that adding these additional probabilistic facts $0.5::q_i$, does not affect the probability distribution on formulas over a vocabulary consisting of only $p_1,\ldots,p_k$. 

In the following proposition, $\calw_{\bar{s}}$ denotes the set of worlds assigning truth values to  propositional variables in $\bar{s}$.

\begin{proposition}\label{prop:notaffect}
    Let $\calp_{\bar{p}}: \calw_{\bar{p}}\longrightarrow [0.0, 1.0]$  be a probability distribution on $\calw_{\bar{p}}$, specified by the first line in~\eqref{eq:PiAp},  $\calp_{\bar{p},\bar{q}}: \calw_{\bar{p},\bar{q}}\longrightarrow [0.0, 1.0]$ be the whole probability distribution specified in the first two lines of~\eqref{eq:PiAp}, and let $\calt$ be an arbitrary theory over vocabulary $\bar{p}$.  Then\break  $\calp_{\bar{p}}\big(\calt\big)=\calp_{\bar{p},\bar{q}}\big(\calt\big)$, where the probability of $\calt$ is defined by~\eqref{eq:probtheory}.
\end{proposition}
\noindent \textbf{Proof}:
According to~\eqref{eq:probtheory}, \[\calp_{\bar{p},\bar{q}}\big(\calt\big)\;\defeq\!\!\!\! \sum_{w\in\calw_{\bar{p},\bar{q}},\; w\models\calt}\!\!\!\!\calp_{\bar{p},\bar{q}}\big(w\big).\]
By~\eqref{eq:PiAp}, the probability of each $q_i\in\bar{q}$ is $\displaystyle\frac{1}{2}$. There are $n$ variables in $\bar{q}$, so   $\displaystyle\calp_{\bar{p},\bar{q}}\big(w\big)=\calp_{\bar{p}}\big(w\big)*\frac{1}{2^n}$. Therefore,
\[\calp_{\bar{p},\bar{q}}\big(\calt\big)\;=\!\!\!\! \sum_{w\in\calw_{\bar{p},\bar{q}},\; w\models\calt}\!\!\!\!\calp_{\bar{p}}\big(w\big)*\frac{1}{2^n}.\]
Every world $w\in \calw_{\bar{p},\bar{q}}$ extends the corresponding world $w'\in \calw_{\bar{p}}$ by  assigning truth values to each variable in $\bar{q}$. Each such world $w'$ has then $2^n$ copies in $\calw_{\bar{p},\bar{q}}$ (each copy accompanied by a different assignment of truth values to variables in $\bar{q}$). Therefore,  
\[
\calp_{\bar{p},\bar{q}}\big(\calt\big) =  \displaystyle 2^n*\!\!\!\!\sum_{w\in\calw_{\bar{p}},\; w\models\calt}\!\!\!\!\calp_{\bar{p}}\big(w\big)*\frac{1}{2^n} \;\;=\!\!\!\! \displaystyle \sum_{w\in\calw_{\bar{p}},\; w\models\calt}\!\!\!\!\calp_{\bar{p}}\big(w\big)\;=\calp_{\bar{p}}\big(\calt\big).
\]
\mbox{}\\[-2.5em]\mbox{}\done\\

\section{Computing Propositional Loss Measures: an~Example}\label{sec:example-prop}

In this example, the theory $\calt_c$ will be used to generate both model-counting-based and proba\-bilistic-based loss measures for a forgetting policy $\bar{p}= \{ecar, jcar\}$.

In Section~\ref{sec:compmodcount}, it was shown how $\calt_c$ could be transformed into a stratified logic program where its probability variables were all assigned the value $0.5$, reflecting a uniform probability on worlds. In this case, $\calp_{0}(\calt_{c}) =0.203125$.
In Section~\ref{sec:compprob}, a probability distribution where $ecar$ and $jcar$ were assigned probabilities, $0.2$ and $0.3$, respectively, resulted in $\calp_{0}(\calt_{c}) =0.3125$.

In the next step, $\nforget{\calt_c; ecar,jcar}$ and $\sforget{\calt_c; ecar,jcar}$ need to be reduced to propositional theories using quantifier elimination. Using results in Section~\ref{sec:forgetting}, forgetting $ecar$ and $jcar$ in theory $\calt_c$, given by formulas~\eqref{eq:cars1}--\eqref{eq:cars3} in Section~\ref{sec:intro}, is defined by:
\begin{align}
    & \nforget{\calt_c; ecar,jcar} \equiv  \exists ecar\exists jcar\big(\calt_c\big);\label{eq:exnforget}\\
    & \sforget{\calt_c; ecar,jcar} \,\equiv  \forall ecar\forall jcar\big(\calt_c\big).\label{eq:exsforget}
\end{align}
Using formulas~\eqref{eq:cars1}--\eqref{eq:cars3} from Section~\ref{sec:intro}, together with~\eqref{eq:exnforget} and \eqref{eq:exsforget}, it can be shown that:\footnote{Here the \dls algorithm~\cite{dls}, together with obvious simplifications of the resulting formula, has been used.}
\begin{itemize}
    \item \nforget{\calt_c; ecar,jcar} is equivalent to:
\begin{equation}\label{eq:execarjcar1}
\!\!\!\!\begin{array}{ll}
  \exists ecar\exists jcar  \big(\!\!\!\! & jcar\implies (car\land reliable\land fcar)\land \\
    & ecar\implies (car\land fast\land  fcar)\land \\
    & fcar\implies (ecar\lor jcar)\;\big). 
\end{array}
\end{equation}
Using~\eqref{eq:exists} from Section~\ref{sec:forgetting}, and some equivalence preserving transformations, one can show  that~\eqref{eq:execarjcar1} is equivalent to the propositional formula:
\begin{equation}\label{eq:execarjcar2}
 fcar\implies ((car\land fast)\lor (car\land reliable)). 
\end{equation}

\item \sforget{\calt_c; ecar,jcar} is equivalent to:
\begin{equation}\label{eq:execarjcar3}
\!\!\!\!\begin{array}{ll}
  \forall ecar\forall jcar  \big( jcar\implies (car\land reliable\land fcar)\big)\land \\
   \forall ecar\forall jcar  \big( ecar\implies (car\land fast\land  fcar\big)\land \\
   \forall ecar\forall jcar  \big( fcar\implies (ecar\lor jcar)\big), 
\end{array}
\end{equation}
i.e, to:
\begin{equation}\label{eq:execarjcar4}
\begin{array}{ll}
   \big(car\land reliable\land fcar\big)\land 
   \big(car\land fast\land  fcar\big)\land 
   \lneg fcar, 
\end{array}
\end{equation}
which simplifies to \false.
\end{itemize}

The next step in the working methodology is to encode $\nforget{\calt_c; ecar,jcar}$ as a stratified logic program in \problog. There is no need to encode $\sforget{\calt_c; ecar,jcar}$, since it is equivalent to \false and its probability $\prob{\sforget{\calt_c; ecar,fcar}} = 0.0$.

\begin{remark}
    Note that this is a limiting case for $\sforget{\calt_c; \bar{p}}$, where the resultant theory is inconsistent. Similarly, but not in this example, the limiting case for $\nforget{\calt_c; \bar{p}}$ would be where the resultant theory is a tautology. In the former case, the forgetting policy is simply too strong, whereas in the latter, it would be too weak. What is important to observe is that the computational techniques proposed can identify these limiting cases, where probabilities of the resultant theories would be $0.0$ and $1.0$, respectively.\done
\end{remark}

\begin{program}[ht]
  \caption{\problog program $\pi^0_{\mbox{\scriptsize\eqref{eq:execarjcar2}}}$ encoding formula \eqref{eq:execarjcar2} equivalent to\\ 
  \hspace*{1.8cm} \nforget{\calt_c; ecar,jcar}\label{prog:example1}}
 $r_{\lneg fcar}\when\ \pneg fcar.$\;
 $r_{car\land fast}\when\ car, fast.$\;
 $r_{car\land reliable}\when\ car, reliable.$\;
 $r_{(car\land fast)\lor (car\land reliable)}\when\ r_{car\land fast}.$\;
 $r_{(car\land fast)\lor (car\land reliable)}\when\ r_{car\land reliable}.$\;
 $r_{fcar\implies((car\land fast)\lor (car\land reliable))}\when\ r_{\lneg fcar}.$\label{line:46}\;
 $r_{fcar\implies((car\land fast)\lor (car\land reliable))}\when\ r_{(car\land fast)\lor(car\land reliable)}.$\label{line:47}
\end{program}

Program~\ref{prog:example1} contains rules encoding formula \eqref{eq:execarjcar2} which is equivalent to the strong forgetting \nforget{\calt_c; ecar,jcar}. In order to compute model-counting-based loss measures, probabilistic facts of the form $0.5::p$ for each propositional variable $p$ occurring in  \nforget{\calt_c; ecar,jcar} are added to Program~\ref{prog:example1}:
\begin{equation}\label{eq:05p}
    0.5\!::\!car.\;  0.5\!::\!reliable.\;\;  0.5\!::\!fast.\;\;   0.5\!::\!fcar.\; ,
\end{equation} 
To determine the probability $\prob{\nforget{\calt_c; ecar,fcar}}$, one sets up a \problog query about the probability of the auxiliary variable $r_{fcar\implies((car\land fast)\lor (car\land reliable))}$, defined in Lines~\ref{line:46}--\ref{line:47} of Program~\ref{prog:example1}. The query returns $\prob{\nforget{\calt_c;ecar,fcar}} = 0.6875$.

In order to compute the probabilistic-based loss measures, the probability distribution is the same as in Program~\ref{prog:tcarprob}. That is, rather than adding specification~\eqref{eq:05p} to Program~\ref{prog:example1}, one should instead add Lines~\ref{line:31}--\ref{line:32} of Program~\ref{prog:tcarprob}. As before, the probability of \nforget{\calt_c; ecar,jcar} is given by the probability of its encoding specified by the auxiliary variable $r_{fcar\implies((car\land fast)\lor (car\land reliable))}$.
 But in fact, this would not change the probability $\prob{\nforget{\calt_c;ecar,jcar}}$.
 Since formula  $\nforget{\calt_c; ecar,jcar}$ does not contain the variables $ecar, jcar$, its probability is still $0.6875$ when computing both $loss^{T}_m$ and $loss^{T}_p$.
 
The values of model counting-based as well probabilistic-based measures of forgetting for $\calt_c$ with  forgetting policy $\bar{p}= \{ecar, jcar\}$ are shown in Table~\ref{tab:results}.

\begin{table}[ht]
    \centering
    \caption{The values of  measures of forgetting $ecar, jcar$ from $\calt_c$.  \label{tab:results}}
    \def\arraystretch{1.3}
    \begin{tabular}{c|c|c||c|c|c} 
    \hline
       $loss^{NC}_m$   &   $loss^{SC}_m$   &  $loss^{T}_m$  & 
         $loss^{NC}_p$   &  $loss^{SC}_p$  &   $loss^{T}_p$\\ \hline
        0.484375  &  0.203125  &  0.6875  &  0.375  &  0.3125  &  0.6875  \\ \hline
    \end{tabular}
\end{table}

Notice that, in this particular case, $loss^{NC}_m$ is greater than $loss^{NC}_p$. That indicates that the inferential strength wrt necessary conditions is lost to a greater extent when we count equally probable models, compared to the case when models get nonequal probabilities. At the same time, as regards reasoning with sufficient conditions, $loss^{SC}_m$ is smaller than $loss^{SC}_p$. In this case, though the same models result from forgetting, the probabilistic measures indicate how the inferential strength changes when real-world probabilities of models are taken into account.

As already discussed, the inferential strength of a theory is strongly related to the number of worlds satisfying the theory, so it is also useful to compare how many worlds are represented by $loss^{NC}_m(\calt_c,ecar,jcar)$ and by $loss^{SC}_m(\calt_c,ecar,jcar)$, i.e., satisfying respectively theories \nforget{\calt_c; ecar,jcar} and \sforget{\calt_c; ecar,jcar}, with the number of worlds satisfying $\calt_c$. The following example illustrates the idea.

Since in theory $\calt_c$ there are $6$ propositional variables, there are $2^6=64$ worlds, so:
 \begin{itemize}
     \item the probability $\prob{\calt_c} = 0.203125$, so there are  $0.203125*64=13$ worlds satisfying $\calt_c$;
     \item the number of worlds represented by $loss^{NC}_m(\calt_c,ecar,jcar)$ is $0.484375*64=31$, being the number of additional worlds satisfying \nforget{\calt_c; ecar,jcar}. One can then expect more formulas valid in all worlds  satisfying $\calt_c$ than  valid in all worlds satisfying \nforget{\calt_c; ecar,jcar}. In consequence, one can expect more formulas that can be inferred from $\calt_c$ than from \nforget{\calt_c; ecar,jcar} what indicates a lost in the inferential strength wrt necessary conditions after strong forgetting;
     \item $loss^{SC}_m(\calt_c,ecar,jcar)=0.203125$, so there are $64*0.203125=13$  worlds lost after applying \sforget{\calt_c; ecar,jcar}. That is, no worlds remain, meaning that the only formulas implying \sforget{\calt_c; ecar,jcar} are those equivalent to false, what illustrates a substantial decrease of the inferential strength wrt sufficient conditions for the theory $\calt_c$. 
 \end{itemize}

In the probabilistic case the situation is analogous, but more probable worlds count more so more strongly affect the loss functions than the less probable ones.

\section{Extending Measures to the 1st-Order Case}\label{sec:fo}

In this section, generalization of the techniques considered to the classical 1st-order logic case are described. 

For the 1st-order language, one assumes finite sets of relation symbols, $\calr$, (1st-order) variables, $\calv_I$, and constants, \calc.  The semantics assumed  for 1st-order logic is standard, assuming that It is  assumed that \calc is the {\em domain of 1st-order interpretations (worlds)} considered.

In addition to propositional connectives, quantifiers $\forall, \exists$ ranging over the domain are allowed.\footnote{Function symbols are avoided, as is standard in rule-based query languages and underlying belief bases~\cite{AHV}.} 
A variable occurrence $x$ is {\em bound} in a formula $A$ when it occurs in the scope of a~quantifier $\forall x$ or $\exists x$. A variable occurrence is {\em free} when the occurrence is not bound. In the rest of this section, $A(\bar{x})$ is used to indicate that $\bar{x}$ are all variables that occur free in $A$. Formula $A$ is {\em closed} when it contains no free variables. 

By an {\em atomic formula} ({\em atom}, for short), one means any formula of the form $r(\bar{z})$, where $r\in\calr$, each  $z$ in $\bar{z}$ is a~variable in $\calv_I$ or a~constant in \calc.  
By a {\em ground atom}, one means an atom containing only constants, if any.

A {\em 1st-order theory (belief base)} is a finite set of closed 1st-order formulas, \calt, understood as a single formula being the conjunction of formulas in \calt.\footnote{The assumption as to the closedness of formulas is used to simplify the presentation. It can be dropped in a~standard manner by assuming an external assignment of constants to free variables.} 
In what follows, it is always assumed that the set of constants (thus the domain) consists of all and only constants occurring in the considered theory. By the {\em set of ground atoms} of theory \calt, $\calg_\calt$, one means all ground atoms with constants occurring in \calt.

Rules of {\em 1st-order stratified logic programs}  have the form:
\begin{equation}\label{eq:prologrulefirst}
 h(\bar{x})\when \pm r_1(\bar{x}_1), \ldots, \pm r_k(\bar{x}_1). 
\end{equation}
where $k\geq 0$, one assumes that $\bar{x},\bar{x}_1, \ldots \bar{x}_k$ are (possibly empty) tuples of variables/constants such that any variable in $\bar{x}$ occurs in at least one $\bar{x}_i$ ($1\leq i\leq k$),  and the stratification condition formulated in Section~\ref{sec:problog} applies.

Similarly, in \problog's probabilistic  facts, annotated disjunctions and queries one can use 1st-order atomic formulas rather than propositional variables (as defined for the propositional case in Section~\ref{sec:problog}).

\begin{remark}\label{rem:dca}
When a particular theory \calt is considered, by restricting the domain to constants occurring in \calt, one always deals with a~finite domain, since the considered theories are finite sets of formulas. That is, a kind of  Domain Closure Axiom is assumed.   \done
\end{remark}

By a {\em world for a 1st-order theory (formula)} \calt one means any assignment of truth values to ground atoms in $\calg_\calt$:
\begin{equation}\label{eq:world-fo}
	w: \calg_\calt\longrightarrow\{\true, \false\}.
\end{equation}
The set of worlds of \calt is denoted by $\calw_\calt$. 

\begin{remark}\label{rem:worlds}
	Observe that worlds defined as in \eqref{eq:world-fo} can be seen as relational databases, where rows in tables contain ground atoms being true in a given world. For example, let the vocabulary contain a unary relation symbol $r$, a binary relation symbol $s$, and constants $a,b$. Let a~world $w$ assign the truth values $w(r(a))=\true$, $w(s(a,b))=w(s(b,b))=\true$, and \false to all other ground atoms. Then $w$ can be seen as a relational database with two tables: $r$ with one row containing $a$, and $s$ with two rows containing $\tuple{a,b}, \tuple{b,b}$. 
	
	Accordingly, the evaluation of formulas in worlds can be seen as evaluation of database queries, expressed by formulas~\cite{AHV}. \done
\end{remark}

As shown in~\cite{dsz-forgetting-aij}, the following theorem holds, extending Theorem~\ref{thm:forget} to the 1st-order case.

\begin{theorem}\label{thm:nforget-fo}
	For arbitrary tuples of relation symbols $\bar{r},\bar{s}$ and a closed 1st-order theory $Th(\bar{r},\bar{s})$,
	\begin{enumerate}
		\item 
		$\nforget{Th(\bar{r},\bar{s}); \bar{r}}  \equiv \exists \bar{r}\,\big(Th(\bar{r},\bar{s})\big).$
		
		\item \mbox{}
		$    \sforget{Th(\bar{r},\bar{s}); \bar{r}}\equiv \forall \bar{r}\big(Th(\bar{r},\bar{s})\big).
		$
  \done
	\end{enumerate}
\end{theorem}

The extension of definitions of probability structures (Definition~\ref{def:probtheory}), model counting-based (Definition~\ref{def:mcmeasures}) and probabilistic-based (Definition~\ref{def:probmeasures}) measures is now straightforward when it is assumed that worlds assign truth values to ground atoms rather than to propositional variables, and that {\em models} are worlds where given theories (formulas) are true. The same symbols, $\lossm{}, \lossp$, will be used to denote these measures. Note that the definitions still satisfy Propositions~\ref{prop:properties},  \ref{prop:propnc} and \ref{prop:subadd} generalized to the 1st-order case. In particular, Point~1, analogous to that of Proposition~\ref{prop:properties}, follows from Equation~\eqref{eq:important}. 

In order to use \problog, transformation of formulas provided in Table~\ref{tab:transform} has to be adjusted to cover the extended logical language, as shown in Table~\ref{tab:transformI}, where rather than auxiliary propositional variables, auxiliary relation symbols are used.

\begin{table}[ht]
\caption{Transformation of an arbitrary 1st-order formula to a~stratified logic program. We specify variables $\bar{x}$ to indicate all free variables appearing in the input formula.
\label{tab:transformI}}
\centering
\begin{tabular}{|c|}\hline\mbox{}\\[-0.5em]
\qquad 
$
\!\!\pi^I_A\! \defeq \left\{\begin{array}{ll}
   \!\!\!\!\begin{array}{l}\emptyset \mbox{\ (the empty program);}\\[-0.7mm]
   r_A\!\defeq\! A \end{array} & \hspace*{-7mm}\mbox{\;\;\;when } A \mbox{ is an atom};\\[-1.5mm] 
   \!- - - - - \\[-1.5mm] 
 \!\!\!\!\begin{array}{l}
        \pi^I_B. \\[-0.7mm]
        r_A(\bar{x})\when \pneg r_B(\bar{x}).  
   \end{array}
      & \hspace*{-3mm} \mbox{when } A=\lneg B(\bar{x});\\[-1.5mm]  
   \!- - - - - \\[-1.5mm] 
   \!\!\!\!\begin{array}{l}
   \pi^I_C.\; \pi^I_D. \\[-0.7mm]
   r_A(\bar{x})\when r_C(\bar{x}), r_D(\bar{x}).\end{array} 
   & \hspace*{-3mm} \mbox{when } A=(C\land D)(\bar{x});\\[-1.5mm] 
   \!- - - - - \\[-1.5mm] 
   \!\!\!\!\begin{array}{l}\pi^I_C.\; \pi^I_D. \\[-0.7mm]
   r_A(\bar{x})\when r_C(\bar{x}).\\[-0.7mm]
   r_A(\bar{x})\when r_D(\bar{x}).\end{array} & \hspace*{-3mm} \mbox{when } A=(C\lor D)(\bar{x});\\[-1.5mm] 
   \!- - - - - \\[-1.5mm] 
   \!\!\!\!\begin{array}{l}\pi^I_C.\; \pi^I_D. \\[-0.7mm]
   r_A(\bar{x})\when \pneg r_C(\bar{x}).\\[-0.7mm]
   r_A(\bar{x})\when r_D(\bar{x}).\end{array} & \hspace*{-3mm} \mbox{when } A=(C\!\implies\! D)(\bar{x});\\[-1.5mm] 
   \!- - - - - \\[-1.5mm] 
   \!\!\!\!\begin{array}{l}\pi^I_C.\; \pi^I_D. \\[-0.7mm]
   r_{A'}(\bar{x})\when \pneg r_C(\bar{x}).\\[-0.7mm]
   r_{A'}(\bar{x})\when r_D(\bar{x}).\\[-0.7mm]
   r_{A''}(\bar{x})\when \pneg r_D(\bar{x}).\\[-0.7mm]
   r_{A''}(\bar{x})\when r_C(\bar{x}).\\[-0.7mm]   
   r_A(\bar{x})\when r_{A'}(\bar{x}), r_{A''}(\bar{x}).
   \end{array} & \hspace*{-3mm} \mbox{when } A=(C\equiv D)(\bar{x});\\[-1.5mm]  
   \!- - - - - \\[-1.5mm] 
   \!\!\!\!\begin{array}{l}
        \pi^I_B. \\[-0.7mm]
        r_A(\bar{x})\when r_B(y,\bar{x}).  
   \end{array}
      & \hspace*{-7mm} \mbox{\;\;\;when } A(\bar{x})\!=\!\exists y B(y,\bar{x});\\[-1.5mm] 
      \!- - - - - \\[-1.5mm] 
 \!\!\!\!\begin{array}{l}
        \pi^I_B. \\[-0.7mm]
        r_{B'}(y,\bar{x})\when\! \pneg\!\, r_{\!B}(y,\bar{x}).\\[-0.7mm]
        r_{A'}(\bar{x})\when r_{B'}(y,\bar{x}).\\[-0.7mm]
        r_A(\bar{x})\when \pneg r_{A'}(\bar{x}).  
   \end{array}
      & \hspace*{-7mm} \mbox{\;\;\;when } A(\bar{x})\!=\!\forall y B(y,\bar{x}).
\end{array}\right.
$\qquad\quad
\\ \qquad\quad\\
\hline
\end{tabular}
\end{table}

The transformation provided in Table~\ref{tab:transformI} extends the one for propositional formula given in Table~\ref{tab:transform} by:
\begin{itemize}
    \item covering predicates and their arguments in clauses for propositional connectives;
    \item dealing with $\exists$ by using a variable $y$ that occurs in the body of the rule $r_A(\bar{x})\when r_B(y,\bar{x})$ and does not occur in its head, in the clause for $A(\bar{x})=\exists y B(y,\bar{x})$, what is obtained by applying a~tautology of classical 1st-order logic, being a standard in logic programming;
    \item dealing with $\forall$ by using the fact that $\forall$ is equivalent to $\lneg\exists \lneg$ in the clause for the case $A(\bar{x})=\forall y B(y,\bar{x})$.

\end{itemize}

As an example, consider a formula $A$, being  $\forall x\exists y (s(x,y,a)\land t(y,b))$, where $x,y$ are variables and $a,b$ are constants. The set of rules representing $A$ is provided in Program~\ref{prog:example2}.

\begin{program}
	\caption{A program $\pi^I_{\forall x\exists y (s(x,y,a)\land t(y,b))}$ encoding the formula\\ 
   \hspace*{1.75cm} $\forall x\exists y (s(x,y,a)\land t(y,b))$.
  \label{prog:example2} }
	\% $\pi^I_{s(x,y,a)}$, $\pi^I_{t(y,b)}$ are empty; $r_{s(x,y,a)}(x,y)\!\defeq\! s(x,y,a)$, $r_{t(y,b)}\!\defeq\! t(y,b)$\;
	$r_{s(x,y,a)\land t(y,b)}(x,y)\when\; s(x,y,a), t(y,b) .$\;
	$r_{\exists y(s(x,y,a)\land t(y,b))}(x)\when\; r_{s(x,y,a)\land t(y,b)}(x,y)$.\;
	$r_{\exists y(s(x,y,a)\land t(y,b))'}(x)\when\; \pneg r_{\exists y(s(x,y,a)\land t(y,b))}(x). $\;
	$r_{\forall x\exists y (s(x,y,a)\land t(y,b))'}()\when\; r_{\exists y(s(x,y,a)\land t(y,b))'}(x).$\;
	$r_{\forall x\exists y (s(x,y,a)\land t(y,b))}()\when\; \pneg r_{\forall x\exists y (s(x,y,a)\land t(y,b))'}().$
\end{program}

\begin{remark}\label{rem:so-fo}
	Essentially, while extending the results to 1st-order logic, one simply switches from propositional variables to ground atoms, and then suitably deals with quantifiers. The resulting rules form stratified logic programs, so remain within the \problog syntax. Though rules for quantifiers are similar to standard transformations such as, e.g., in~\cite{lloydtopor}, the use of auxiliary relation symbols allows one to properly deal with negation, by using it in a stratified manner which is not guaranteed by the other cited transformations. Also, while the worst-case length of the formula resulting from Lloyd-Topor transformation may be exponential in the size of input formula,  the proposed transformation guarantees that it is linearly bounded.\done
\end{remark}

Similarly to the transformation given in Table~\ref{tab:transform}, for the transformation given in Table~\ref{tab:transformI}, the following propositions and lemmas apply. 

\begin{proposition}\label{prop:linear-fo}
	For every 1st-order formula $A$, the size of $\pi^I_A$ is liner in the size of $A$.\done
\end{proposition}

\begin{proposition}\label{prop:determines-fo}\mbox{}
	\begin{enumerate}
		\item Let  $A$ be a  1st-order formula. Then each world $w\in\calw_A$ determines the values of auxiliary variables in $\pi^I_A$. 
		\item Given a world $w\in\calw_A$, the truth values of auxiliary relation symbols in $\pi^I_A$ can be determined in deterministic polynomial time in the size of $A$.\done
	\end{enumerate}
\end{proposition}

\noindent
Notice that the polynomial time complexity mentioned in Point~2 of Proposition~\ref{prop:determines-fo} reflects the complexity of 1st-order queries~\cite{AHV}: as indicated in Remark~\ref{rem:worlds}, worlds can be seen as relational databases, and formulas -- as queries to such databases.

\begin{lemma}\label{lemma:strat-fo}
	For every 1st-order formula $A$, $\pi^I_A$ is a~stratified logic program.\done
\end{lemma}

\begin{lemma}\label{lemma:equiv-fo}
	Let $\pi^I_A$ occur in the program $\Pi_A^m$. Then for every subformula $B$ of $A$ occurring in the construction of $\pi^I_A$, we have that $r_B$ is equivalent to $B$.\done
\end{lemma}

The construction of the \problog program $\Pi^m_A$ given in Equation~\eqref{eq:PiAm} is adjusted to the 1st-order case by replacing probabilistic propositional variables by all ground atoms. That is, if $A(\bar{x})$ is a 1st-order formula then $\Pi_A^m$ becomes:
\begin{equation}\label{eq:PiAmI}
	\Pi_A^m\defeq\left|   \begin{array}{l}
		0.5::r_1(\bar{c}_{11}). \ldots 0.5::r_1(\bar{c}_{1m_1}).\\
		\ldots\\
		0.5::r_l(\bar{c}_{l1}). \ldots 0.5::r_l(\bar{c}_{lm_l}).\\ 
		\pi^I_A(\bar{x}).
	\end{array}\right.
\end{equation}
where $r_1, \ldots r_l$ are all relation symbols in $\calr$, for $1\leq i\leq l$, $m_i$ is the arity of relation $r_i$, and $r_1(\bar{c}_{11}), \ldots r_1(\bar{c}_{1m_1}),\ldots r_l(\bar{c}_{l1}), r_l(\bar{c}_{lm_l})$ are all ground atoms with constants occurring in $A(\bar{x})$.

The following theorem, analogous to Theorem~\ref{thm:probA}, applies.

\begin{theorem}\label{thm:probA-fo}
	For every 1st-order formula $A$, the probability of $A$ (in the sense of definition given by~\eqref{eq:probth}) is equal to the probability of $r_A$ given by the \problog program $\Pi^m_A$, defined by~\eqref{eq:PiAmI}. \done
\end{theorem}

The construction of the 1st-order version of $\Pi^p_A$ adjusts the construction \eqref{eq:PiAp} as follows:
\begin{equation}\label{eq:PiApI}
  \Pi_A^p\defeq\left|  \begin{array}{l}
    \mbox{\problog\ specification of } \calp_\calw 
    \mbox{ over selected ground atoms specified}
     \mbox{ in~\eqref{eq:PiAmI}}\\
    0.5::a_1.\; \ldots\; 0.5::a_n. \mbox{ where $a_1, \ldots a_n$ are all other ground atoms}  \\ 
   \pi^I_A.
\end{array}\right.
\end{equation}

\noindent For the above construction we also have a proposition analogous to Proposition~\ref{prop:notaffect}.

Table~\ref{tab:comparison} provides a comparison of transformations given in Tables~\ref{tab:transform} and \ref{tab:transformI} and the transformations of~\cite{tseitin} and \cite{lloydtopor} from the point of view of properties used in this paper. The meaning of rows is the following:
\begin{itemize}
    \item `Satisfiability' indicates whether a given transformation preserves satisfiability (i.e.,\break whether the formula and its transformation are equisatisfiable);
    \item `Equivalence' indicates whether a given transformation preserves equivalence (in the sense of Lemma~\ref{lemma:equiv});\footnote{In particular, assuming complete worlds to make sure that the negation `\pneg' coincides with classical negation.}
    \item `\#Models' indicates whether a given transformation preserves the number of models (i.e., whether the formula and its transformation have the same number of models); 
    \item `Size' -- the (worst case) size of the resulting formula wrt the size of the input formula; 
    \item `1st-order' -- whether the technique is applicable to 1st-order logic; 
    \item `Stratified' -- whether the result is guaranteed to be stratified wrt negation.
\end{itemize}

\begin{table}[ht]\small
    \centering
        \caption{A comparison of properties of formula transformations considered in the paper.}
    \label{tab:comparison}
    \begin{tabular}{l|c|c|c|c}
        & Tseitin~\cite{tseitin} &  Llloyd-Topor~\cite{lloydtopor} & Table~\ref{tab:transform} & Table~\ref{tab:transformI}\\ \hline
      Satisfiability   &  $+$ & $+$ & $+$ & $+$\\
      Equivalence   & $-$ & $-$ & $+$ & $+$ \\
      \#Models & $+$ & $-$ & $+$ & $+$ \\
      Size & linear & exponential & linear & linear\\
      1st-order & $-$ & $+$ & $-$ & $+$ \\
      Stratified &  $-$ & $-$ & $+$ & $+$\\ \hline
    \end{tabular}
\end{table}

\noindent Observe that only the transformations introduced in this paper enjoy the desired properties:  Table~\ref{tab:transform} for propositional logic and Table~\ref{tab:transformI} for 1st-order logic.

\section{Computing 1st-order Loss Measures: an Example}\label{sec:example-fo}

The following example, based on~\cite{dsz-forgetting-aij}, illustrates the use of 1st-order loss measures.

Consider the following belief base, where $ms(x)$ stands for ``person $x$ has mild symptoms of a disease'', $ss(x)$  -- for ``person $x$ has severe symptoms of the disease'', $h(x)$ -- for ``$x$ should stay home'',  $t(x)$ -- for \mbox{``$x$ needs a test for the disease''}, and $ich(x)$ -- for ``$x$ should immediately consult a health care provider'':
	\begin{equation}\label{eq:test}
	\!	Th(ms, h, t, ss, ich)\!=\!\Big\{\forall x \Big(ms(x)\implies \big(h(x)\land t(x)\big) \Big),
		\forall x\Big(\big(ss(x)\lor t(x)\big)\implies ich(x)\big) \Big)\Big\}.
	\end{equation}
When a test is not available, it is useful to forget about it, so one can consider strong and weak forgetting operators \nforget{Th\big(ms, h, t, ss, ich\big); t)} and \sforget{Th\big(ms, h, t, ss, ich\big); t)}. The corresponding derivations are provided in~\cite{dsz-forgetting-aij}:
		\begin{equation}\label{eq:testacknc}
			\begin{array}{l}
				\nforget{Th\big(ms, h, t, ss, ich\big); t)}\equiv \\
				 \qquad\exists t\Big(\forall x \Big(ms(x)\implies \big(h(x)\land t(x)\big) \Big)\land \forall x\Big(\big(ss(x)\lor t(x)\big)\implies ich(x)\Big)\Big)\equiv\\
				 \qquad \forall x \Big(ms(x)\implies h(x)\Big)\land 
				 \forall x\Big(\big(ss(x)\lor ms(x)\big)\implies ich(x) \Big).
			\end{array}
		\end{equation}
			\begin{equation}\label{eq:testacksc}   
			\begin{array}{ll}
				\sforget{Th\big(ms, h, t, ss, ich\big); t)}\equiv \\
				 \qquad\forall t\Big(\forall x \Big(ms(x)\implies \big(h(x)\land t(x)\big) \Big)\land \forall x\Big(\big(ss(x)\lor t(x)\big)\implies ich(x)\Big)\Big)\equiv\\
				 \qquad\forall x\Big(\lneg ms(x)\Big)\land \forall x\Big(ich(x)\Big).
			\end{array}
		\end{equation}

\noindent As discussed in~\cite{dsz-forgetting-aij},
\begin{itemize}
	\item The result (last line of~\eqref{eq:testacknc}) of \nforget{Th\big(ms, h, t, ss, ich\big); t)} indicates that a person with mild symptoms should stay at home. If, under the circumstances, it is not possible to test for the disease, then the severe \textit{or} mild symptoms should suffice to immediately consult the health care provider.
	\item The result (last line of~\eqref{eq:testacksc}) of \sforget{Th\big(ms, h, t, ss, ich\big); t)} indicates that without access to a test one can only make sure that the first formula of~\eqref{eq:test} is true by assuring  that $\forall x\Big(\lneg ms(x)\Big)$ is true. The second formula of~\eqref{eq:test} can only be guaranteed when $\forall x\Big(ich(x)\Big)$ is true.
\end{itemize}

Let us now show how can one compute loss measures related to the above forgetting policy using \problog.  
As an example, the \problog Program~\ref{prog:example6} encodes formulas in belief base \eqref{eq:test}, where some obvious abbreviations are used to avoid separate encoding of conjunctions, as in Lines~\ref{line:prog61}, and~\ref{line:prog62}. The source codes for programs encoding formulas, given in the last lines of \eqref{eq:testacknc}, \eqref{eq:testacksc}, are provided in Program~\ref{prog:example6} in the Appendix.

\begin{program}
	\caption{\label{prog:example6}\problog program $\pi^I_{\eqref{eq:test}}$ encoding the formulas in \eqref{eq:test}.} 
	\% $\pi^I_{ms(x)}, \pi^I_{h(x)}, \pi^I_{t(x)}, \pi^I_{ss(x)}, \pi^I_{ich(x)}$ are empty\; 
	\% $r_{ms(x)}\defeq ms(x), r_{h(x)}\defeq h(x), r_{t(x)}\defeq t(x), r_{ss(x)}\defeq ss(x), r_{ich(x)}\defeq ich(x)$\;
	$r_{ms(x)\implies (h(x)\land t(x))}(x)\when\; \pneg ms(x)$.\;
	$r_{ms(x)\implies (h(x)\land t(x))}(x)\when\; h(x), t(x)$.\label{line:prog61}\hspace*{35.3mm}\% $ms(x)\implies (h(x)\land t(x))$\;
	$r_{(ms(x)\implies (h(x)\land t(x))'}(x)\when\; \pneg r_{ms(x)\implies (h(x)\land t(x)}(x)$.\hspace*{12.4mm}  \% $\lneg(ms(x)\implies (h(x)\land t(x)))$\;
	$r_{\forall x(ms(x)\implies (h(x)\land t(x))'}()\when\;  r_{(ms(x)\implies (h(x)\land t(x))'}(x)$.\hspace*{11.4mm}  \% $\exists x(\lneg(ms(x)\implies (h(x)\land t(x))))$\;
	$r_{\forall x(ms(x)\implies (h(x)\land t(x))}()\when\; \pneg r_{\forall x(ms(x)\implies (h(x)\land t(x))'}()$.\quad\ \ \ \ \ \% the first formula in~\eqref{eq:test}\;\vspace*{0.5em}
	$r_{(ss(x)\lor t(x))\implies ich(x)}(x)\when\; \pneg ss(x), \pneg t(x)$.\label{line:prog62}\;
	$r_{(ss(x)\lor t(x))\implies ich(x)}(x)\when\; ich(x)$.   \hspace*{38.7mm}\% $(ss(x)\lor t(x))\implies ich(x)$\;

	$r_{((ss(x)\lor t(x))\implies ich(x))'}(x)\when\; \pneg r_{(ss(x)\lor t(x))\implies ich(x)}(x)$. \hspace*{8mm}\% $\lneg((ss(x)\lor t(x))\implies ich(x))$\;
	$r_{\forall x((ss(x)\lor t(x))\implies ich(x))'}()\when\; \pneg r_{((ss(x)\lor t(x))\implies ich(x))'}(x)$. \hspace*{3.5mm}\% $\exists x(\lneg((ss(x)\lor t(x))\implies ich(x)))$\;
	$r_{\forall x((ss(x)\lor t(x))\implies ich(x))}()\when\; \pneg r_{\forall x((ss(x)\lor t(x))\implies ich(x))'}()$.\hspace*{4.3mm}\% the second formula in~\eqref{eq:test}\;\vspace*{0.5em}
	$r_{bel\_base}\when\; r_{\forall x(ms(x)\implies (h(x)\land t(x))}(), r_{\forall x((ss(x)\lor t(x))\implies ich(x))}()$.\ \ \% the conjunction of formulas  \hspace*{9.18cm} \% in belief base~\eqref{eq:test}
\end{program}

Assume that one wants to apply the theory to a case of $Eve$. Table~\ref{tab:results-fo} show the values of measures, when the theory is applied to $Eve$, where probabilistic measures are obtained by assuming the probability of $ich$ being $0.8$. That is, in the latter case it is assumed that, statistically, she contacts health care providers $8$ times per $10$ similar cases. The results show that the loss of inferential strength is relatively small both as shown by the model counting-based measures as well as by probabilistic-based ones.

\begin{table}[ht]
	\centering
	\caption{The values of  measures of forgetting $t$ from $Th\big(ms, h, t, ss, ich\big)$.  \label{tab:results-fo}}
	\def\arraystretch{1.3}
	\begin{tabular}{c|c|c||c|c|c} 
		\hline
		$loss^{NC}_m$   &   $loss^{SC}_m$   &  $loss^{T}_m$  & 
		  $loss^{NC}_p$   &  $loss^{SC}_p$  &   $loss^{T}_p$\\ \hline
		0.0625   &  0.0625  &  0.125  &  0.1  &  0.1  &  0.2  \\ \hline
	\end{tabular}
\end{table}

\noindent When one changes the probability of $ich$  to $0.3$, probabilistic measures of $\nlossp{}$,  $\slossp{}$,  respectively are $0.0375$, $0.0375$. This indicates the type of influence of $ich$ on the inferential strength: the smaller its probability is, the smaller loss we have.  

It is also interesting how the probability of $t$ influences the measures. Of course, after forgetting, it has no influence on the probabilities of theories resulting from forgetting $t$. However, it may change the probability of the original theory. For example, when the probability of $t$ is 0.1 then the probability of the original theory is $0.1575$. After changing the probability of $t$ to $0.9$, the probability of the original theory is $0.2175$. Since the probability of strong forgetting is $0.225$ and of the weak forgetting is $0.15$, we have that $loss^{NC}_p$ changes from $0.0675$ to $0.0075$, and $loss^{SC}_p$ -- from $0.0075$ to $0.0675$. Note that we substantiated the theory with just one person `Eve', so there are $2^5=32$ worlds. Taking that into account together with the relatively small difference $0.0675-0.0075=0.06$, one can estimate that the probability of $t$ does not seriously affect the inferential strength after its forgetting.

\section{Related Work}\label{sec:relwork}

Strong (standard) forgetting, \nforget{}, has been introduced in the foundational paper~\cite{Lin94forgetit}, where model theoretical definitions and analysis of properties of the standard $forget()$ operator are provided. The paper~\cite{Lin94forgetit} opened a~research subarea, summarized, e.g., in~\cite{DitmarschHLM09} or more recently, e.g., in~\cite{dsz-forgetting-aij,Eiter-kern-isberner}.  Numerous applications of forgetting include~\cite{BeierleKSBR19,Del-PintoS19,Delgrande17,Eiter-kern-isberner,goncalves,DitmarschHLM09,WangZZZ14,ZHANG2006739,ZhaoSchmidt16b,ZhaoS17}. In summary, while the topic of forgetting  has gained considerable attention in knowledge representation,  measures of forgetting have not been proposed in the literature. Notice that  the weak forgetting operator \sforget{}, introduced in~\cite{dsz-forgetting-aij}, plays a substantial role in definitions of measures $\slossm{}, \tlossm{}, \slossp{}$ and $\tlossp{}$.

The model counting problem \#SAT as well as \#SAT-solvers have been extensively investigated  \cite{ChakrabortyMV21,fichte,GomesSS21,SoosM19}. The methods and algorithms for \#SAT  provide a solid alternative for implementing model counting-based measures. In particular projected model counting, such as that considered in~\cite{GebserKS09,LagniezLM20,LagniezM19}, could be used for computing the measures \nlossm{} since variables are projected out using existential quantification  which directly corresponds to standard forgetting (see Point 1 of Theorem~\ref{thm:forget}). 

\problog has been chosen as a~uniform framework for computing both model counting-based as well as probabilistic measures. This, in particular, has called for a new formula transformation, as shown in Table~\ref{tab:transform}. Our transformation is inspired by Tseitin's transformation of arbitrary propositional formulas into the Conjunctive Normal Form (see, e.g.,~\cite{KuiterKSTS22,LagniezLM20,tseitin}). Though Tseitin's transformation preserves satisfiability, it does not preserve equivalence, nor does it lead to stratified clauses. On the other hand, our transformation preserves equivalence in the sense formulated in Lemma~\ref{lemma:equiv} and allows its use together with probabilistic logic programming languages that use stratified negation, including \problog. For 1st-order quantifiers, the transformation shown in Table~\ref{tab:transformI} uses a rather standard technique, similar to the well-known transformation of~\cite{lloydtopor}. However, the use of auxiliary relation symbols again allows us to obtain stratified logic programs. On the other hand, the transformations of~\cite{lloydtopor} and alike could not be directly applied here as they generally do not ensure stratification nor the linear size of the resulting formula. In fact, the worst-case length of the formula resulting from Lloyd-Topor transformation may be exponential in the size of input formula. For a comparison of Tseitin's, Llloyd-Topor and our transformations from the point of view of properties needed in our paper, see Table~\ref{tab:comparison}.

In the past decades there has been a fundamental interest in combining logic with probabilities, in particular in the form of probabilistic programming languages which could also be used for computing measures such as those proposed in our paper. Such languages include ~{\sc Clp(Bn)}~\cite{SantosCosta2008}, {\sc Cp}~\cite{VennekensDB06}, {\sc Icl}~\cite{Poole08}, {\sc LP}$^{\mbox{\scriptsize\sc Mln}}$~\cite{lpmln},  {\sc Prism}~\cite{SatoK97},  {\sc ProPPR}~\cite{WangMC13}. For books surveying related approaches see~\cite{deraedt-book,Pfeffer2016,riguzzi}. Our choice of \problog is motivated by the  rich and extensive research around it,  both addressing theory and applications, as well as by its efficient implementation. 

In order to deal with 2nd-order quantifiers, which are inevitably associated with forgetting, one can use a variety of techniques -- see, e.g., \cite{gss} and, for the context of forgetting,~\cite{dsz-forgetting-aij}. For eliminating 2nd-order quantifiers in this paper, the \dls algorithm~\cite{dls} has been used.

The proposed measures considered in this paper are relatively independent of a particular formalism since one only requires algorithms for counting models or probabilities over them.  For example, forgetting is particularly useful in rule-based languages, when one simplifies a~belief base to improve querying performance. This is especially useful for forgetting in Answer Set Programs~\cite{goncalves,WangZZZ14,ZHANG2006739}, where the corresponding entailment tasks, centering around necessary conditions, are typically intractable. Our approach can be  adjusted to this framework using algorithms for counting ASP models, like those investigated in~\cite{AzizCMS15}.

\section{Conclusions}\label{sec:concl}

Two types of loss measures have been proposed, one model counting-based and the other probabilistic-based, to  measure the losses in inferential strength for theories where different forgetting policies and operators are used. The model counting-based approach is based on an underlying uniform probability distribution for worlds, whereas  the probabilistic-based approach is based on the use of arbitrary probability distributions on worlds. The former can be seen as a non-weighted approach and the latter as a weighted approach to measuring inferential strength and the losses ensued through the use of forgetting. A computational framework based on \problog is proposed that allows analysis of any propositional or closed 1st-order theory with finite domains relative to different forgetting policies. 

The idea of measuring the inferential strength of forgetting is novel. To the best of our knowledge, neither these nor other quantitative measures of forgetting have been considered in the literature so far. The techniques shown in the paper can be applied to numerous examples reported in the literature on forgetting. As discussed in the paper, quantitative measures can also be used to automate the choice of symbols to forget, e.g., in abstracting or query optimization.

The proposed approach is very general (see also Remark~\ref{rem:general}) and can be adopted to other logics or rule-based systems used in AI or other application areas. Good examples are description logics/ontology languages and (non-monotonic) Answer Set Programming.  Given that one has a (weighted) model counter for the considered logic, the pragmatic computational tool is also there. \problog has been used for reasons explained in the paper. A neat side effect of using \problog is that one obtains a probabilistic logic programming language with queries expressible by arbitrary formulas of the considered background logic. The methods proposed here allow for using all 1st-order formulas as probabilistic logic programming queries. Equally important is the fact that computing such queries requires \problog programs of linear size wrt the computed queries.

The framework proposed should have applications beyond forgetting. For instance, the example in Section~\ref{sec:intro} has been used in~\cite{NayakL95} as a means of modeling abstraction through the use of forgetting where one abstracts from $ecar, jcar$. The loss measures proposed in this paper should be useful as one criterion for  determining degrees of abstraction. Another broad area of potential applications of the introduced measures is knowledge compilation and theory approximation~\cite{CadoliD97,DarwicheM24,kautzselmanK96}. For example, in~\cite{kautzselmanK96}, knowledge compilation and theory approximation is used to compile propositional theories into Horn formulas that approximate the original information. The approximations bound the original theory from below and above in terms of inferential strength. The measures we propose can be used to assess relative strength of such approximations/compilations. Earlier in the paper, it was pointed out that model counts can be derived directly from the model-counting-based approach to loss. It would be interesting to empirically compare the use of \problog and the techniques proposed here with existing model counting approaches as mentioned in Section~\ref{sec:relwork}. 

New techniques have also been proposed for transforming arbitrary propositional and 1st-order formulas into stratified logic programs, not necessarily \problog. As shown in Table~\ref{tab:comparison}, only the new transformations proposed in this paper are linear in the size of the input formula and at the same time preserve satisfiability, equivalence, stratifiability and the number of models. 
This makes the transformations efficient and applicable to any stratified logic programming or rule-based language. In particular, this applies to  \problog, where preserving equivalence, stratifiability and the number of models is essential for correctly computing the probabilities of queries expressed by arbitrary propositional or 1st-order formulas.

It should finally be mentioned that all \problog programs specified in this paper are accessible for experimentation and verification using a~\problog web interface.\footnote{For the convenience of the reader, \problog sources, in ready to copy and paste form are included in the Appendix.}

\section*{Appendix A: \problog Programs}

The following programs can be tested using the \problog standalone interpreter or its web interface mentioned in the end of Section~\ref{sec:problog}.

\subsection*{Program~\ref{prog:tcar}}

The following program  computes 
the probability of theory $\calt_c$ expressed by~\eqref{eq:cars1}--\eqref{eq:cars3}.

{\footnotesize
	\begin{verbatim}
		0.5::car.  0.5::reliable.  0.5::fast.
		0.5::fcar. 0.5::jcar.      0.5::ecar.
		
		r :- car, reliable, fcar.
		r :- \+jcar.   % r represents (1)
		s :- car, fast, fcar.
		s :- \+ecar.   % s represents (2)
		u:- jcar.
		u:- ecar.  
		u:- \+ fcar.   % u represents (3)
		t :- r, s, u.  % t represents (1)&(2)&(3)
		query(t).
	\end{verbatim}
}

\subsection*{Program~\ref{prog:example}}

The following program computes the probability of the formula $\big(q\equiv (p\land\lneg q\land s)\big)$.

{\footnotesize
	\begin{verbatim}
  0.5::p. 0.5::q. 0.5::s.  

  r_not_q :- \+ q.
  r_p_and_not_q_and_s :- p, r_not_q, s.
  r_q_equiv_p_and_not_q_and_s_1 :-  \+ q.
  r_q_equiv_p_and_not_q_and_s_1 :-  r_p_and_not_q_and_s.
  r_q_equiv_p_and_not_q_and_s_2 :-  \+ r_p_and_not_q_and_s.
  r_q_equiv_p_and_not_q_and_s_2 :-  q. 
  r_q_equiv_p_and_not_q_and_s :-  r_q_equiv_p_and_not_q_and_s_1, 
                                  r_q_equiv_p_and_not_q_and_s_2.
  query(r_q_equiv_p_and_not_q_and_s).
	\end{verbatim}
}

\subsection*{Program~\ref{prog:tcarprob}}

The following program  computes 
the probability of theory $\calt_c$ expressed by~\eqref{eq:cars1}--\eqref{eq:cars3} when\break $\prob{ecar}=0.2$ and $\prob{jcar}=0.3$, and the choices of $jcar$ and $ecar$ exclude each other.

{\footnotesize
	\begin{verbatim}
		0.2::ecar; 0.3::jcar.
		0.5::car.  0.5::reliable.  
		0.5::fast. 0.5::fcar.  
		
		r :- car, reliable, fcar.
		r :- \+jcar.   % r represents (1)
		s :- car, fast, fcar.
		s :- \+ecar.   % s represents (2)
		u:- jcar.
		u:- ecar.  
		u:- \+ fcar.   % u represents (3)
		t :- r, s, u.  % t represents (1)&(2)&(3)		
		query(t).
	\end{verbatim}
}

\subsection*{Program~\ref{prog:example1}}

The following program computes the probability of the formula~\eqref{eq:execarjcar2} equivalent to\break $\nforget{\calt_c; ecar,jcar}$. To compute the probability of~\eqref{eq:execarjcar2} when $\prob{ecar}=0.2$ and\break  $\prob{jcar}=0.3$, and $ecar, jcar$ exclude each other, it suffices to replace the first line by: {\footnotesize\texttt{0.2::ecar; 0.3::jcar.}}

{\footnotesize
	\begin{verbatim}
		0.5::ecar. 0.5::jcar.
		0.5::car.  0.5::reliable.  
		0.5::fast. 0.5::fcar.       
		
		r_not_fcar :- \+ fcar.
		r_car_and_fast :- car, fast.
		r_car_and_reliable :- car, reliable.
		r_car_and_fast_or_car_and_reliable :- r_car_and_fast.
		r_car_and_fast_or_car_and_reliable :- r_car_and_reliable.
		r_fcar_impl_car_and_fast_or_car_and_reliable :-r_not_fcar.
		r_fcar_impl_car_and_fast_or_car_and_reliable 
		                   :- r_car_and_fast_or_car_and_reliable.		
		query(r_fcar_impl_car_and_fast_or_car_and_reliable).
	\end{verbatim}
}

\mbox{}\\[-4.5em]
\subsection*{Program~\ref{prog:example2}}

The following program  computes 
the probability of formula $\forall x\exists y (s(x,y,a)\land t(y,b))$, assuming that the underlying domain consists of $a, b$ (expressed by $dom(a), dom(b)$).

{\footnotesize
	\begin{verbatim}
		0.5::s(X,Y,Z):- dom(X), dom(Y), dom(Z).
		0.5::t(X,Y):- dom(X), dom(Y).
		
		dom(a). dom(b). 
		
		r_sxya_and_tyb(X,Y) :- s(X,Y,a), t(Y,b).
		r_exists_y_sxya_and_tyb(X) :-  r_sxya_and_tyb(X,Y).
		r_exists_y_sxya_and_tyb_1(X) :-  \+ r_exists_y_sxya_and_tyb(X). 
		r_forall_x_exists_y_sxya_and_tyb_1:-  r_exists_y_sxya_and_tyb_1(X).
		r_forall_x_exists_y_sxya_and_tyb :- \+ r_forall_x_exists_y_sxya_and_tyb_1.	
		query(r_forall_x_exists_y_sxya_and_tyb).
	\end{verbatim}
} 

\mbox{}\\[-4.5em]
\subsection*{Program~\ref{prog:example6}}

The following program computes probabilities needed for loss values shown in Table~\ref{tab:results-fo}:
\begin{itemize}
	\item  the probability of belief base~\eqref{eq:test}: being the value of variable {\footnotesize\texttt{r\_bel\_base}}, according to rules  specified in Program~\ref{prog:example6};
	\item  the probability of \nforget{Th\big(ms, h, t, ss, ich\big); t)} given by the last line of~\eqref{eq:testacknc}: being the value of variable {\small\texttt{r\_F\_NC\_Th\_ms\_h\_t\_ss\_ich}};
	\item  the probability of \sforget{Th\big(ms, h, t, ss, ich\big); t)} given by the last line of~\eqref{eq:testacksc}: being the value of variable {\small\texttt{r\_F\_SC\_Th\_ms\_h\_t\_ss\_ich}}. 
\end{itemize}
The values of probabilistic measures shown in Table~\ref{tab:results-fo} are obtained by changing the probabilistic fact {\small\texttt{0.5::ich(eve)} into  {\small\texttt{0.8::ich(eve)}.

Notice that in the first line we use an annotated disjunction since mild and severe symptoms exclude each other.

{\footnotesize
\begin{verbatim}
  0.5::ms(eve); 0.5::ss(eve). 
  0.5::h(eve). 0.5::t(eve).  0.8::ich(eve).

% the first formula in the belief base:
  r_ms_x_implies_h_x_and_t_x(X) :- \+ ms(X).
  r_ms_x_implies_h_x_and_t_x(X) :- h(X), t(X).
  r_ms_x_implies_h_x_and_t_x_1(X):- \+ r_ms_x_implies_h_x_and_t_x(X).
  r_forall_x_ms_x_implies_h_x_and_t_x_1 :- r_ms_x_implies_h_x_and_t_x_1(X).
  r_forall_x_ms_x_implies_h_x_and_t_x :- 
                        \+ r_forall_x_ms_x_implies_h_x_and_t_x_1.

% the second formula in the belief base:                        
  r_ss_x_or_t_x_implies_ich_x(X) :- \+ ss(X), \+ t(X).
  r_ss_x_or_t_x_implies_ich_x(X) :- ich(X).
  r_ss_x_or_t_x_implies_ich_x_1(X):- \+ r_ss_x_or_t_x_implies_ich_x(X).
  r_forall_x_ss_x_or_t_x_implies_ich_x_1 :- 
                        r_ss_x_or_t_x_implies_ich_x_1(X). 
  r_forall_x_ss_x_or_t_x_implies_ich_x :- 
                        \+ r_forall_x_ss_x_or_t_x_implies_ich_x_1.

% conjunction of formulas in the belief base:
  r_bel_base :- r_forall_x_ms_x_implies_h_x_and_t_x, 
                r_forall_x_ss_x_or_t_x_implies_ich_x. 
  query(r_bel_base).    

% the first conjunct of F^NC:
  r_ms_x_implies_h_x(X) :- \+ ms(X).
  r_ms_x_implies_h_x(X) :- h(X).
  r_ms_x_implies_h_x_1(X) :- \+ r_ms_x_implies_h_x(X).
  r_forall_x_ms_x_implies_h_x_1 :- r_ms_x_implies_h_x_1(X).
  r_forall_x_ms_x_implies_h_x :- \+ r_forall_x_ms_x_implies_h_x_1.
                        
% the second conjunct of F^NC:   
  r_ss_x_or_ms_x_implies_ich_x(X) :- \+ ss(X), \+ ms(X).
  r_ss_x_or_ms_x_implies_ich_x(X) :- ich(X).
  r_ss_x_or_ms_x_implies_ich_x_1(X):- \+ r_ss_x_or_ms_x_implies_ich_x(X).
  r_forall_x_ss_x_or_ms_x_implies_ich_x_1 :- 
                        r_ss_x_or_ms_x_implies_ich_x_1(X). 
  r_forall_x_ss_x_or_ms_x_implies_ich_x :- 
                        \+ r_forall_x_ss_x_or_ms_x_implies_ich_x_1.
                         
% F^NC(Th(ms, h, t, ss, ich); t):                         
  r_F_NC_Th_ms_h_t_ss_ich :- r_forall_x_ms_x_implies_h_x,
                             r_forall_x_ss_x_or_ms_x_implies_ich_x.                  
  query(r_F_NC_Th_ms_h_t_ss_ich). 

% the first conjunct of F^SC:
  r_forall_x_not_ms_x_1(X) :-  ms(X).
  r_forall_x_not_ms_x :- \+ r_forall_x_not_ms_x_1(X). 
                           
% the second conjunct of F^SC:                         
  r_not_ich_x(X) :- \+ ich(X).
  r_forall_x_ich_x_1 :- r_not_ich_x(X).
  r_forall_x_ich_x :- \+ r_forall_x_ich_x_1. 

% F^SC(Th(ms, h, t, ss, ich); t):
  r_F_SC_Th_ms_h_t_ss_ich :- r_forall_x_not_ms_x, r_forall_x_ich_x.                 
  query(r_F_SC_Th_ms_h_t_ss_ich). 
	\end{verbatim}
}

\section*{Acknowledgments}

This research has been supported by  a research grant from the ELLIIT Network Organization for Information and Communication Technology, Sweden. The first author has also been supported by a research grant from Mahasarakham University, Thailand.

\small
\let\oldthebibliography\thebibliography
\let\endoldthebibliography\endthebibliography
\renewenvironment{thebibliography}[1]{
	\begin{oldthebibliography}{#1}
		\setlength{\itemsep}{1pt}
		\setlength{\parskip}{0pt}
	}
	{
	\end{oldthebibliography}
}

\end{document}